%% file: vqa_main.tex
\pdfoutput=1
\documentclass[10pt,journal,letterpaper,twoside,compsoc]{IEEEtran}

\usepackage{times}

\usepackage{epsfig}
\usepackage{graphicx}
\usepackage[belowskip=-10pt,aboveskip=2pt,font=small]{caption}
\usepackage{dblfloatfix}

\usepackage{amsmath}
\usepackage{amssymb}

\usepackage{multirow}
\usepackage{rotating}
\usepackage{booktabs}
\usepackage{slashbox} 
\usepackage{algpseudocode}
\usepackage{algorithm}

\usepackage{enumerate}
\usepackage[olditem,oldenum]{paralist}

\usepackage{mysymbols}

\usepackage[pagebackref=true,breaklinks=true,letterpaper=true,colorlinks,bookmarks=false,citecolor=blue,linkcolor=green]{hyperref}

\usepackage{color}
\usepackage{comment}
\input{space_saver}

\definecolor{orange}{RGB}{225, 90, 0}
\definecolor{teal}{RGB}{5, 210, 150}
\definecolor{yellow}{RGB}{220, 210, 10}
\definecolor{purple}{RGB}{100, 0, 205}

\newcommand{\change}{\textcolor{black}}

\newcommand{\changenew}{\textcolor{black}}

\begin{document}

\title{VQA: Visual Question Answering\\
\large{\url{www.visualqa.org}}}

\author{Aishwarya Agrawal$^{*}$, Jiasen Lu$^{*}$, Stanislaw Antol$^{*}$, \\
Margaret Mitchell, C.~Lawrence~Zitnick, Dhruv Batra, Devi Parikh
\IEEEcompsocitemizethanks{
\IEEEcompsocthanksitem $^*$The first three authors contributed equally.
\IEEEcompsocthanksitem A. Agrawal, J. Lu and S. Antol are with Virginia Tech.
\IEEEcompsocthanksitem M. Mitchell is with Microsoft Research, Redmond. \IEEEcompsocthanksitem C.~L.~Zitnick is with Facebook AI Research.
\IEEEcompsocthanksitem D. Batra and D. Parikh are with Georgia Institute of Technology.}
}



\IEEEcompsoctitleabstractindextext{\begin{abstract}

We propose the task of \emph{free-form} and \emph{open-ended} Visual Question Answering (VQA). Given an image and a natural language question about the image, the task is to provide an accurate natural language answer. Mirroring real-world scenarios, such as helping the visually impaired, both the questions and answers are open-ended. Visual questions selectively target different areas of an image, including background details and underlying context. As a result, a system that succeeds at VQA typically needs a more detailed understanding of the image and complex reasoning than a system producing generic image captions.
Moreover, VQA is amenable to automatic evaluation, since many open-ended answers contain only a few words or a closed set of answers
that can be provided in a multiple-choice format. We provide a dataset containing $\sim$0.25M images, $\sim$0.76M questions, and $\sim$10M answers (\url{www.visualqa.org}), and discuss the information it provides. Numerous baselines and methods for VQA are provided and compared with human performance. Our VQA demo is available on CloudCV (\url{http://cloudcv.org/vqa}). 
\end{abstract}}

\maketitle

\input{introduction}
\input{related_work}
\input{collection}
\input{analysis}
\input{baselines}
\input{challenge}
\input{discussion}

\input{vqa_supplement_v2}


\clearpage
{\footnotesize
\bibliographystyle{ieee}
\bibliography{vqa_main}
}

\end{document}

%% file: space_saver.tex
\belowdisplayskip \abovedisplayskip

\newlength{\sectionReduceTop}
\newlength{\sectionReduceBot}
\newlength{\subsectionReduceTop}
\newlength{\subsectionReduceBot}
\newlength{\abstractReduceTop}
\newlength{\abstractReduceBot}
\newlength{\captionReduceTop}
\newlength{\captionReduceBot}
\newlength{\subsubsectionReduceTop}
\newlength{\subsubsectionReduceBot}

\newlength{\eqnReduceTop}
\newlength{\eqnReduceBot}

\newlength{\horSkip}
\newlength{\verSkip}

\newlength{\figureHeight}

%% file: introduction.tex
\section{Introduction}
\label{sec:intro}

We are witnessing a renewed excitement in multi-discipline Artificial Intelligence (AI) research problems. In particular, research in image and video captioning that combines Computer Vision (CV),
Natural Language Processing (NLP), and Knowledge Representation \& Reasoning (KR) has dramatically increased in the past year \cite{captioning_msr,captioning_xinlei,captioning_berkeley,captioning_baidu_ucla,
captioning_toronto,captioning_stanford,captioning_google}. Part of this excitement stems from a belief that multi-discipline tasks like image captioning are a step towards solving AI. However, the current state of the art demonstrates that a coarse scene-level understanding of an image paired with word $n$-gram statistics suffices to generate reasonable image captions, which suggests image captioning may not be as ``AI-complete'' as desired.

What makes for a compelling ``AI-complete'' task?
We believe that in order to spawn the next generation of AI algorithms,
an ideal task should
\begin{inparaenum}[(i)]
\item require \emph{multi-modal knowledge} beyond a single sub-domain (such as CV) and
\item have a well-defined \emph{quantitative evaluation metric} to track progress.
\end{inparaenum}
For some tasks, such as image captioning, automatic evaluation is still a difficult and
open research problem \cite{cider, elliott2014comparing, hodosh2013framing}.

\begin{figure}[t]
\includegraphics[width=1\linewidth]{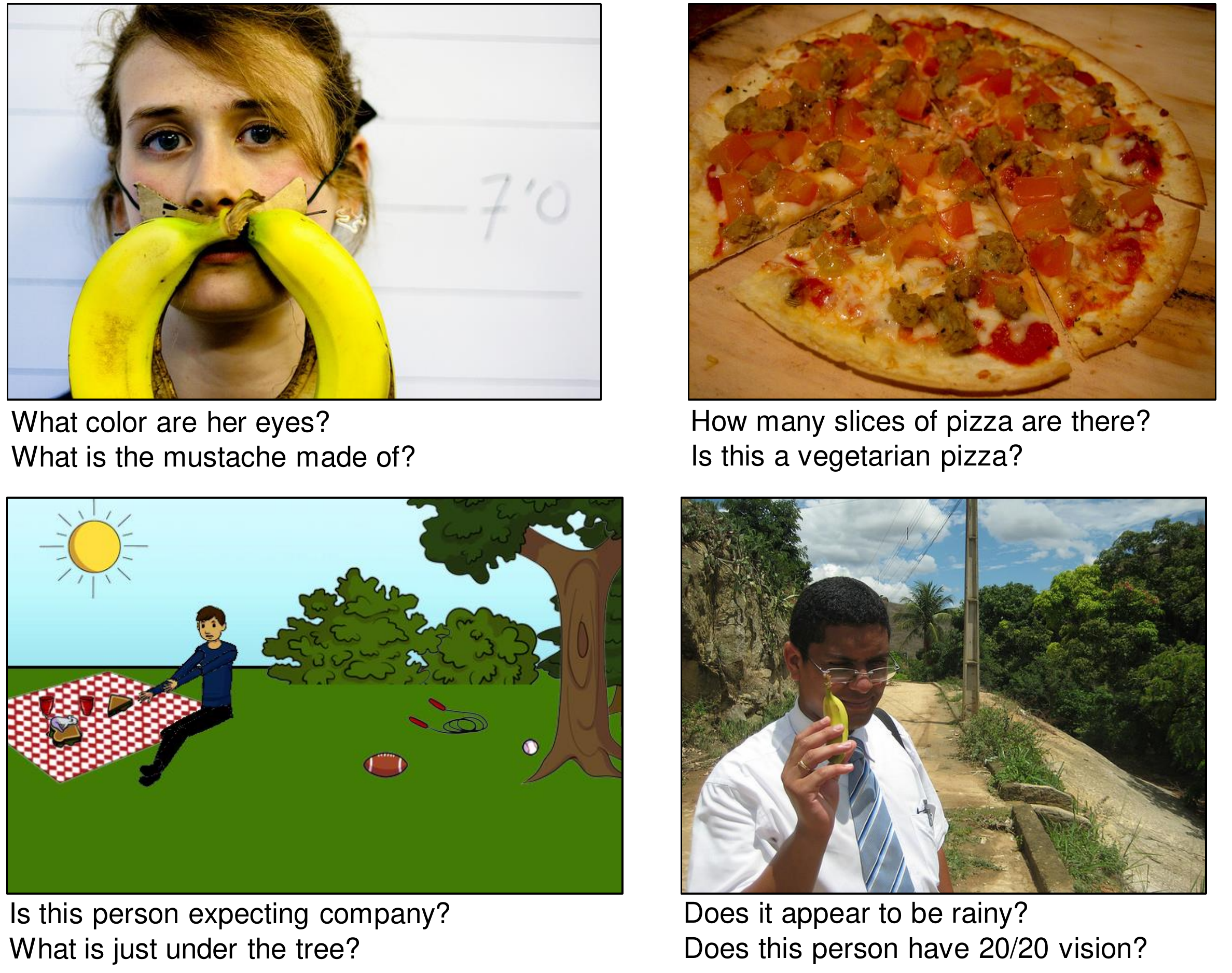}
\caption{Examples of free-form, open-ended questions collected for images via Amazon Mechanical Turk.
Note that commonsense knowledge is needed along with a visual understanding of the scene
to answer many questions.}
\label{fig:teaser}
\end{figure}

In this paper, we introduce the task of \emph{free-form} and \emph{open-ended}
Visual Question Answering (VQA).
A VQA system takes as input an image and a free-form, open-ended, natural-language question
about the image and produces a natural-language answer as the output.
This goal-driven task is applicable to scenarios encountered when visually-impaired
users~\cite{vizwiz} or intelligence analysts actively elicit visual information.
Example questions are shown in \figref{fig:teaser}.

Open-ended questions require a potentially vast
set of AI capabilities to answer --
fine-grained recognition (\eg, ``What kind of cheese is on the pizza?''),
object detection (\eg, ``How many bikes are there?''),
activity recognition (\eg, ``Is this man crying?''),
knowledge base reasoning (\eg, ``Is this a vegetarian pizza?''),
and commonsense reasoning (\eg, ``Does this person have 20/20 vision?", ``Is this person expecting company?'').
VQA~\cite{geman,fritz,SongChun_video_queries,vizwiz} is also amenable to automatic quantitative evaluation, making it possible to effectively track progress on this task.
While the answer to many questions is simply ``yes'' or ``no'', the process for determining a correct answer is typically far from trivial (e.g.~in \figref{fig:teaser}, ``Does this person have 20/20 vision?'').
 Moreover, since questions about images often tend to seek specific information, simple one-to-three word answers are sufficient for many questions. In such scenarios, we can easily evaluate a proposed algorithm by the number of questions it answers correctly.
In this paper, we present both an open-ended answering task and a multiple-choice
task \cite{richardson2013mctest,FITB_VP}. Unlike the open-ended task that requires a
free-form response, the multiple-choice task only requires an algorithm to pick from a predefined
list of possible answers.  


We present a large dataset that contains 204,721 images from the MS COCO
dataset \cite{coco} and a newly created abstract scene dataset \cite{ZitnickCVPR2013,Antol2014}
that contains 50,000 scenes. The MS COCO dataset has images depicting diverse and complex scenes that are effective at eliciting compelling and diverse questions. We collected a new dataset of ``realistic'' abstract scenes to enable research focused only on the high-level reasoning required for VQA by removing the need to parse real images. Three questions were collected for each image or scene. Each question was answered by ten subjects along with their confidence. The dataset contains over 760K questions with around 10M answers.

While the use of open-ended questions offers many benefits, it is still useful to understand the types of questions that are being asked and which types various algorithms may be
good at answering. To this end, we analyze the types of questions asked and the types of answers provided.
Through several visualizations, we demonstrate the astonishing diversity of the questions asked. We also explore how the information content of questions and their answers differs from image captions.
For baselines, we offer several approaches that use a combination of both text and state-of-the-art
visual features~\cite{AlexNet}. As part of the VQA initiative, we will organize an annual challenge and
associated workshop to discuss state-of-the-art methods and best practices.

VQA poses a rich set of challenges, many of which have been viewed as the holy grail of
automatic image understanding and AI in general. However, it includes as building blocks
several components that the CV, NLP, and KR \cite{NEEL,NEIL,Cyc,ConceptNet,Freebase} communities have made
significant progress on during the past few decades. VQA provides an attractive balance
between pushing the state of the art, while being accessible enough for the communities
to start making progress on the task.


%% file: related_work.tex
\section{Related Work}
\label{sec:related}
\textbf{VQA Efforts.}
Several recent papers have begun to study visual question answering~\cite{geman,fritz,SongChun_video_queries,vizwiz}.
However, unlike our work, these are fairly restricted (sometimes synthetic) settings with small datasets.
For instance,~\cite{fritz} only considers
questions whose answers come from a predefined closed world of
16 basic colors or 894 object categories.
\cite{geman}
also considers questions generated from templates from a fixed vocabulary of objects, attributes,
relationships between objects, \etc.
In contrast, our proposed task involves \emph{open-ended}, \emph{free-form}
questions and answers provided by humans.
Our goal is to increase the diversity of knowledge and
kinds of reasoning needed to provide correct answers.
Critical to
achieving success on this more difficult and unconstrained task, our VQA dataset is \emph{two orders of magnitude} larger than \cite{geman,fritz}
($>$250,000 \vs 2,591 and 1,449 images respectively). The proposed VQA task has connections to other related work:  ~\cite{SongChun_video_queries} has studied joint parsing of videos and corresponding text to answer queries on two datasets containing 15 video clips each. ~\cite{vizwiz} uses crowdsourced workers to answer questions about visual content asked by visually-impaired users.
In concurrent work, \cite{Malinowski_2015_ICCV} proposed combining an LSTM for the question 
with a CNN for the image to generate an answer. In their model, the LSTM question representation is conditioned on the CNN image features at each time step, and the final LSTM hidden state is used to sequentially decode the answer phrase. In contrast, the model developed in this paper explores ``late fusion'' -- \ie, the LSTM question representation and the CNN image features are computed independently, \emph{fused} via an element-wise multiplication, and then passed through fully-connected layers to generate a softmax distribution over output answer classes.
\cite{Lin_2015_CVPR} generates abstract scenes to capture visual common sense relevant to answering (purely textual) fill-in-the-blank and visual paraphrasing questions. \cite{Sadeghi_2015_CVPR} and \cite{Vendantam_2015_ICCV} use visual information to assess the plausibility of common sense assertions. \cite{Yu_2015_ICCV} introduced a dataset of 10k images and prompted captions that describe specific aspects of a scene (\eg, individual objects, what will happen next). 
Concurrent with our work, \cite{baiduVQA} collected questions \& answers in Chinese (later translated to English by humans) for COCO images. \cite{Ren_2015_NIPS} automatically generated four types of questions (object, count, color, location) using COCO captions.



\textbf{Text-based Q\&A}
is a well studied problem in the NLP and text processing communities (recent examples
being~\cite{zettlemoyer_kdd14,zettleymoyer_acl13,weston_qa,richardson2013mctest}).
Other related textual tasks include sentence
completion (\eg,~\cite{richardson2013mctest} with multiple-choice answers).
These approaches provide inspiration for VQA techniques.
One key concern in text is the \emph{grounding} of questions.
For instance, \cite{weston_qa} synthesized textual descriptions and QA-pairs
grounded in a simulation of actors and objects in a fixed set of locations.
VQA is naturally grounded in images -- requiring the understanding of both
text (questions) and vision (images).
Our questions are generated by humans, making the need for commonsense
knowledge and complex reasoning more essential.

\textbf{Describing Visual Content.}
Related to VQA are the tasks of image tagging~\cite{deng,AlexNet}, image
captioning~\cite{babytalk,FarhadiSentencesECCV2010,MitchellEtAl12,captioning_xinlei,captioning_msr,captioning_google,captioning_berkeley,captioning_stanford,captioning_baidu_ucla,captioning_toronto} and video captioning~\cite{video,youtube2text}, where words or sentences are
generated to describe visual content. While these tasks require both visual and semantic knowledge,
captions can often be non-specific (e.g., observed by \cite{captioning_google}). 
The questions in VQA require detailed specific information about the image for which generic
image captions are of little use \cite{vizwiz}.

\textbf{Other Vision+Language Tasks.} Several recent papers have explored tasks at the intersection of vision and language
that are easier to evaluate than image captioning,
such as
coreference resolution~\cite{nounCoref2014,peopleCoref2014}
or generating referring expressions~\cite{referit,mitchell2013generating}  
for a particular object in an image that would allow a human to identify
which object is being referred to (\eg, ``the one in a red shirt'',
``the dog on the left'').
While task-driven and concrete, a limited set of visual concepts
(\eg, color, location) tend to be captured by referring expressions. As we demonstrate, a
richer variety of visual concepts emerge from visual questions and their answers.

%% file: collection.tex
\section{VQA Dataset Collection}
\label{sec:dataset}
\begin{figure*}[t]
\centering
\includegraphics[width=1\linewidth]{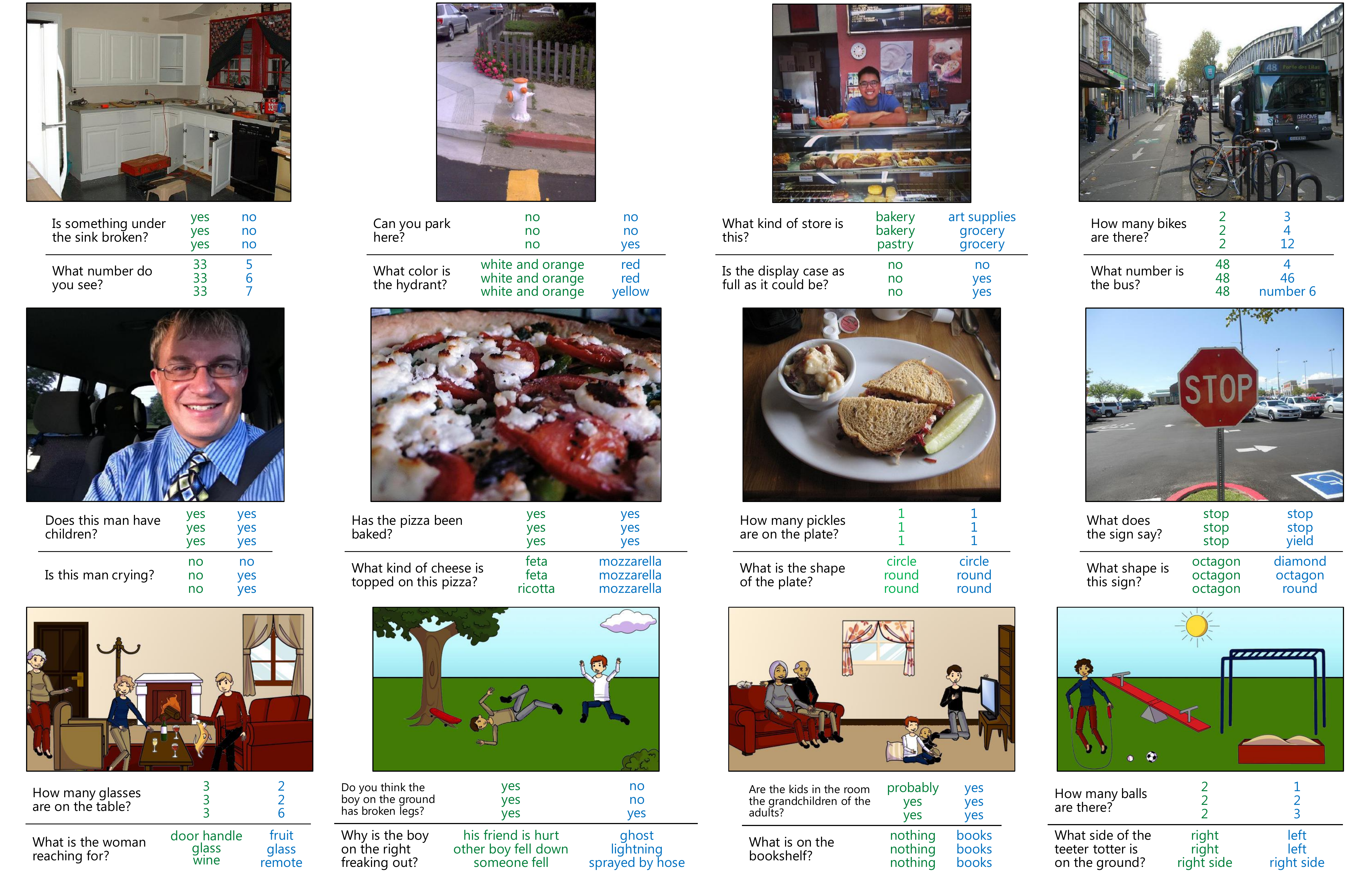}
\caption{Examples of questions (black), (a subset of the) answers given when looking at the image (green), and answers given when not looking at the image (blue) for numerous representative examples of the dataset. See the appendix for more examples.}
\label{fig:qualResults}
\end{figure*}
We now describe the Visual Question Answering (VQA) dataset. We begin by
describing the real images and abstract scenes used to collect the questions. Next, we describe
our process of collecting questions and their corresponding answers. Analysis of the questions
and answers gathered as well as baselines' \& methods' results are provided in following sections.

\textbf{Real Images.}
We use the 123,287 training and validation images and 81,434 test images from the newly-released Microsoft
Common Objects in Context (MS COCO)~\cite{coco} dataset. The MS COCO dataset was gathered to find images containing multiple objects and rich contextual information. Given the
visual complexity of these images, they are well-suited for our VQA task. The more diverse our
collection of images, the more diverse, comprehensive, and interesting the resultant set of
questions and their answers.

\textbf{Abstract Scenes.}
The VQA task with real images requires the use of complex and often noisy
visual recognizers. To attract researchers interested in exploring the high-level reasoning
required for VQA, but not the low-level vision tasks, we create a new abstract scenes
dataset \cite{Antol2014,ZitnickCVPR2013,ZitnickICCV2013,VedantamPAMI2015} containing 50K scenes.
The dataset contains 20 ``paperdoll'' human models \cite{Antol2014} spanning genders, races,
and ages with 8 different expressions. The limbs are adjustable to allow
for continuous pose variations. The clipart may be used to depict both
indoor and outdoor scenes. The set contains over 100 objects and 31 animals in
various poses. The use of this clipart enables the creation of more realistic scenes (see bottom row of \figref{fig:qualResults}) that more closely mirror
real images than previous papers \cite{ZitnickCVPR2013,ZitnickICCV2013,VedantamPAMI2015}.
See the appendix
for the user interface, additional details, and examples. 

\textbf{Splits.}
For real images, we follow the same train/val/test split strategy as the MC COCO dataset~\cite{coco} 
(including test-dev, test-standard, test-challenge, test-reserve). For the VQA challenge (see section \ref{sec:challenge}), test-dev is used for debugging and validation experiments and allows for unlimited submission to the evaluation server. Test-standard is the `default' test data for the VQA competition. When comparing to the state of the art (e.g., in papers), results should be reported on test-standard. Test-standard is also used to maintain a public leaderboard that is updated upon submission. Test-reserve is used to protect against possible overfitting. If there are substantial differences between a method's scores on test-standard and test-reserve, this raises a red-flag and prompts further investigation. Results on test-reserve are not publicly revealed. Finally, test-challenge is used to determine the winners of the challenge.

For abstract scenes, we created splits for standardization, separating the scenes into 20K/10K/20K for train/val/test splits, respectively. There are no subsplits (test-dev, test-standard, test-challenge, test-reserve) for abstract scenes.

\textbf{Captions.}
The MS COCO dataset~\cite{coco,capeval2015} already contains five single-sentence captions for all images.
We also collected five single-captions for all abstract scenes using the same user interface\footnote{\url{https://github.com/tylin/coco-ui}} for collection.

\textbf{Questions.}
Collecting interesting, diverse, and well-posed questions is a significant challenge.
Many simple questions may only require low-level computer vision knowledge,
such as ``What color is the cat?'' or ``How many chairs are present in the scene?''.
However, we also want questions that require commonsense knowledge about the scene,
such as ``What sound does the pictured animal make?''. Importantly, questions should also \emph{require} the image to correctly answer and not be answerable using just commonsense information, \eg, in~\figref{fig:teaser}, ``What is the mustache made of?''.  By having a wide variety of
question types and difficulty, we may be able to measure the continual progress of both
visual understanding and commonsense reasoning.

We tested and evaluated a number of user interfaces for collecting
such ``interesting'' questions. 
Specifically, we ran pilot studies asking human subjects to ask questions about a given image that they believe
a ``toddler'', ``alien'', or ``smart robot'' would have trouble answering.
We found the ``smart robot'' interface to elicit the most interesting and diverse questions. 
As shown in the appendix,
our final interface stated:
\begin{center}
\fbox{
\parbox{0.45\textwidth}
{``\emph{We have built a smart robot. It understands a lot about images. It can
recognize and name all the objects, it knows where the objects are, it can recognize the scene
(\eg, kitchen, beach), people's expressions and poses, and properties of objects (\eg, color of
objects, their texture). Your task is to stump this smart robot!}

\emph{Ask a question about this scene that this smart robot probably can not answer, but any human can easily answer while looking at the scene in the image.}''
}
}
\end{center}
To bias against
generic image-independent questions,
subjects were instructed to ask questions that \emph{require} the image to answer. 

The same user interface was used for both the real images and abstract scenes.
In total, three questions from unique workers were gathered for each image/scene.
When writing a question, the subjects were shown the previous questions already asked for that image to
increase the question diversity. In total, the dataset contains over
$\sim$0.76M questions.

\textbf{Answers.}
Open-ended questions result in a diverse set of possible answers.
For many questions, a simple ``yes'' or ``no'' response is sufficient. However, other questions
may require a short phrase. Multiple different answers may also be correct. For instance, the
answers ``white'', ``tan'', or ``off-white'' may all be correct answers to the same question.
Human subjects may also disagree on the ``correct'' answer, \eg, some saying ``yes'' while
others say ``no''. To handle these discrepancies, we gather \emph{10 answers for each
question from unique workers}, while also ensuring that the worker answering a question did not ask it.
We ask the subjects to provide answers that are ``a brief phrase and not a complete sentence.
Respond matter-of-factly and avoid using conversational language or inserting your opinion.''
In addition to answering the questions, the subjects were asked ``Do you think you were
able to answer the question correctly?'' and given the choices of ``no'', ``maybe'', and ``yes''. See the appendix for more details about the user interface to collect answers.
See \secref{sec:analysis} for an analysis of the answers provided.

For testing, we offer two modalities for answering the questions: (i) \textbf{open-ended} and (ii) \textbf{multiple-choice}.

For the open-ended task, the generated answers are evaluated 
using the following accuracy metric:
\begin{equation*}
\text{accuracy} = \min(\frac{\text{\# humans that provided that answer}}{3},1)
\end{equation*}
\ie, an answer is deemed 100\% accurate if at least 3 workers provided that exact answer.\footnote{In order 
to be consistent with `human accuracies' reported in \secref{sec:analysis}, machine accuracies are  
averaged over all ${10 \choose 9}$ sets of human annotators}
Before comparison, all responses are made lowercase, numbers converted to digits,
and punctuation \& articles removed. We avoid using soft metrics
such as Word2Vec \cite{word2vec}, since they often group together 
words that we wish to distinguish, such as ``left'' and ``right''. We also avoid using evaluation metrics from machine translation such as BLEU and ROUGE because such metrics are typically applicable and reliable for sentences containing multiple words. In VQA, most answers (89.32\%) are single word; thus there no high-order n-gram matches between predicted answers and ground-truth answers, and low-order n-gram matches degenerate to exact-string matching. Moreover, these automatic metrics such as BLEU and ROUGE have been found to poorly correlate with human judgement for tasks such as image caption evaluation \cite{DBLP:journals/corr/ChenFLVGDZ15}.

For multiple-choice task, 18 candidate answers are created for each question. As with the open-ended task,
the accuracy of a chosen option is computed based on the number of human subjects who provided 
that answer (divided by 3 and clipped at 1). We generate a candidate set of correct
and incorrect answers from four sets of answers:
\textbf{Correct:} The most common (out of ten) correct answer.
\textbf{Plausible:} To generate incorrect, but still plausible answers we ask
three subjects to answer the questions without seeing the image. See the appendix for more details about the user interface to collect these answers. If three unique answers
are not found, we gather additional answers from nearest neighbor questions
using a bag-of-words model. The use of these answers helps ensure the image,
and not just commonsense knowledge, is necessary to answer the question.
\textbf{Popular:} These are the 10 most popular answers. For instance, these are ``yes'',
``no'', ``2'', ``1'', ``white'', ``3'', ``red'', ``blue'', ``4'', ``green'' for real images.
The inclusion of the most popular answers makes it more difficult for algorithms to infer the type of
question from the set of answers provided, \ie, learning that it is a ``yes or no'' question just because
``yes'' and ``no'' are present in the answers.
\textbf{Random:} Correct answers from random questions in the dataset.
To generate a total of 18 candidate answers, we first find the union of the correct, plausible, and popular answers. 
We include random answers until 18 unique answers are found.
The order of the answers is randomized. Example multiple choice questions are in the appendix.

Note that all 18 candidate answers are unique. But since 10 different subjects answered every question, it is possible that more than one of those 10 answers be present in the 18 choices. In such cases, according to the accuracy metric, multiple options could have a non-zero accuracy.

%% file: analysis.tex
\section{VQA Dataset Analysis}
\label{sec:analysis}
\begin{figure*}[t]
\centering
\includegraphics[width=1\linewidth]{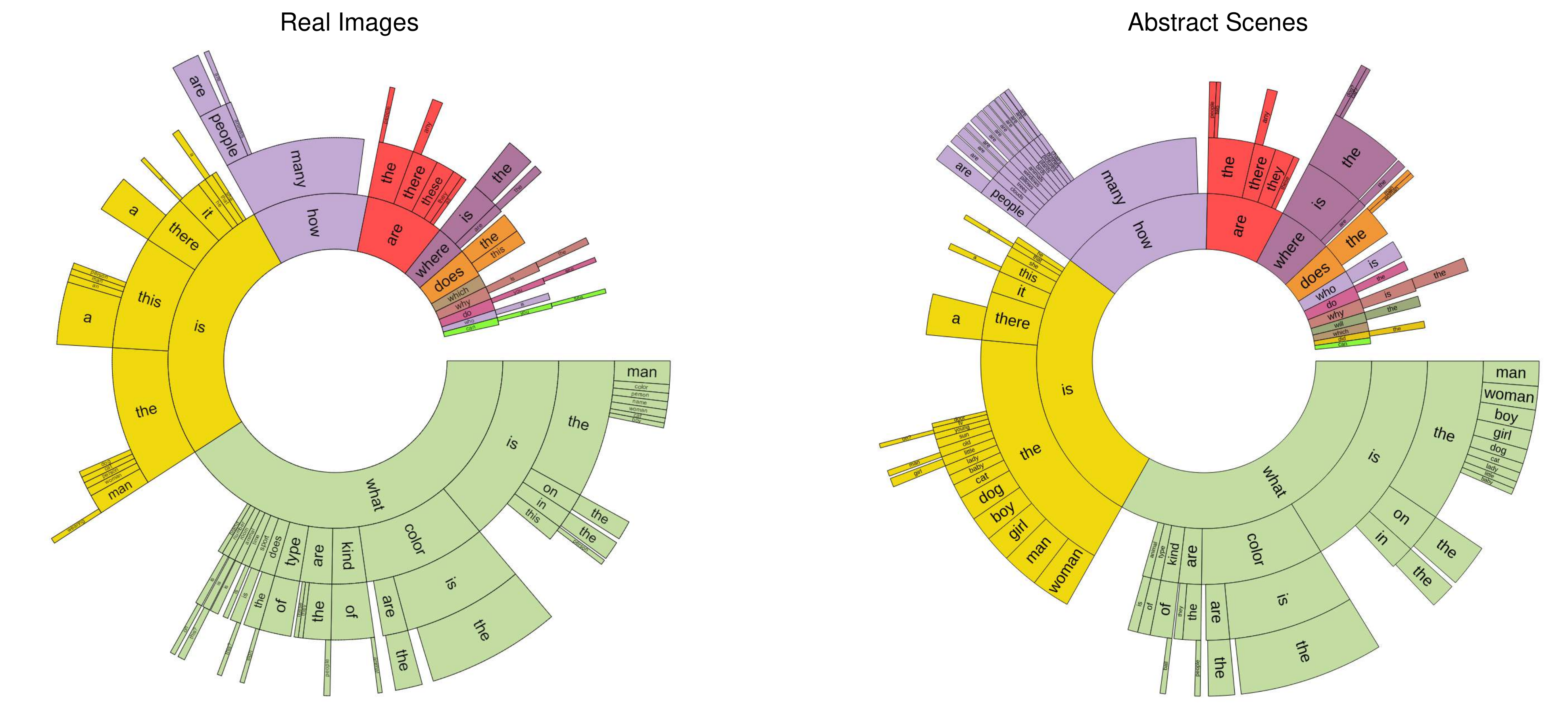}
\caption{Distribution of questions by their first four words for a random sample of 60K questions for real images (left) and all questions for abstract scenes (right). The ordering of the words starts towards the center and radiates outwards. The arc length is proportional to the number of questions containing the word. White areas are words with contributions too small to show. }
\label{fig:QuesCluster}
\end{figure*}

In this section, we provide an analysis of the questions and answers in the VQA train dataset.
To gain an understanding of the types of questions asked and answers provided, we visualize
the distribution of question types and answers. We also explore how often the questions may
be answered without the image using just commonsense information. Finally, we analyze whether
the information contained in an image caption is sufficient to answer the questions.

The dataset includes 614,163 questions 
and 7,984,119 answers (including answers provided by workers with and without 
looking at the image) 
for 204,721 images from the MS COCO dataset~\cite{coco} and 150,000 questions with 1,950,000 answers for $50,000$ abstract scenes.


\subsection{Questions}

\textbf{Types of Question.}
Given the structure of questions generated in the English language,
we can cluster questions into different types based on the words that start the question.
\figref{fig:QuesCluster} shows the distribution of questions based on the first four
words of the questions for both the real images (left) and abstract scenes (right).
Interestingly, the distribution of questions is quite similar for both real images and abstract scenes.
This helps demonstrate that the type of questions elicited by the abstract scenes is similar to
those elicited by the real images. There exists a surprising variety of question types,
including ``What is$\ldots$'', ``Is there$\ldots$'', ``How many$\ldots$'', and ``Does the$\ldots$''.
Quantitatively, the percentage of questions for different types is shown in \tableref{tab:typeacc}. Several example questions and answers are shown in \figref{fig:qualResults}.
A particularly interesting type of question is the ``What is$\ldots$'' questions, since they have a
diverse set of possible answers. See the appendix for visualizations for ``What is$\ldots$'' questions.

\begin{figure}[t]
\centering
\includegraphics[width=1\linewidth]{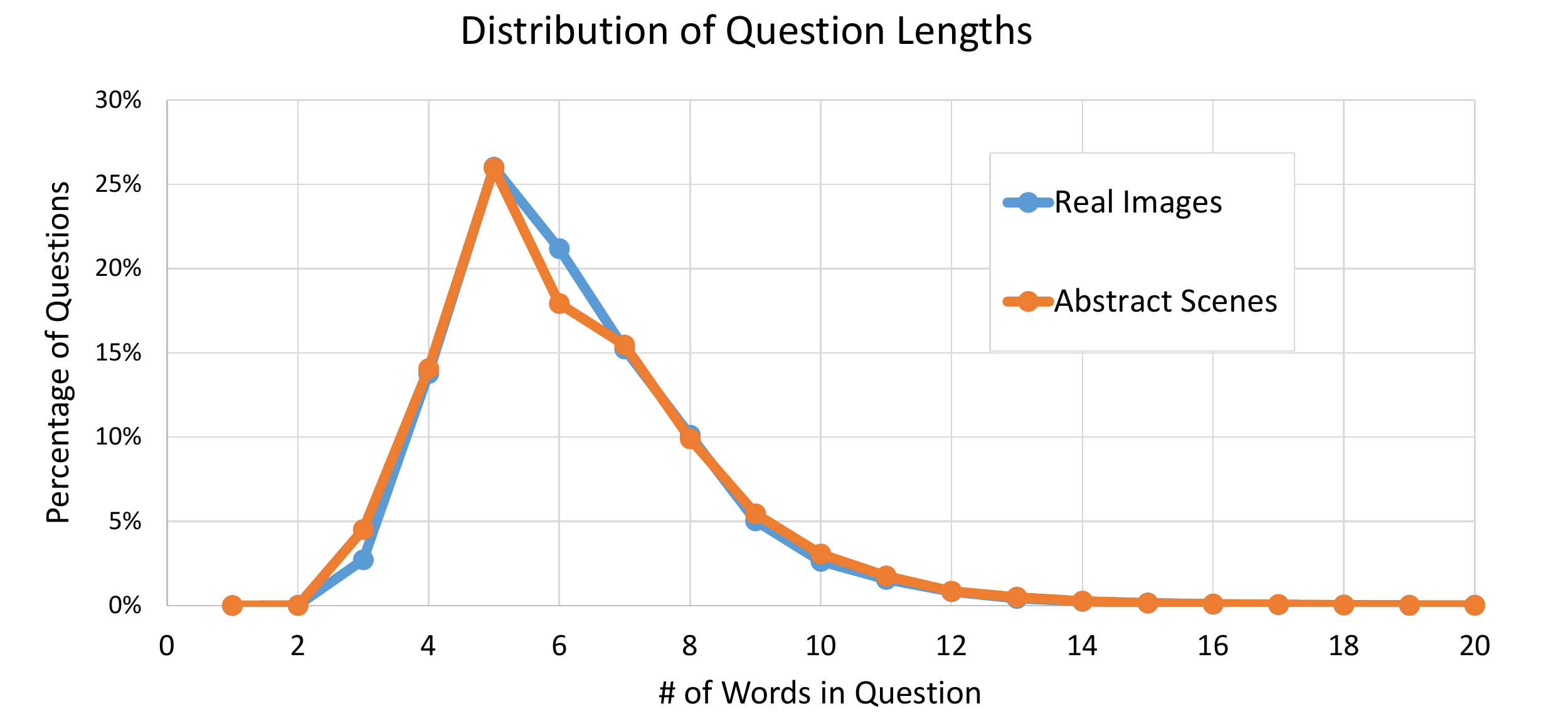}
\caption{Percentage of questions with different word lengths for real images and abstract scenes.}
\label{fig:QuesLen}
\end{figure}

\textbf{Lengths.}
\figref{fig:QuesLen} shows the distribution of question lengths.
We see that most questions range from four to ten words.

\begin{figure*}
\centering
\includegraphics[width=1\linewidth]{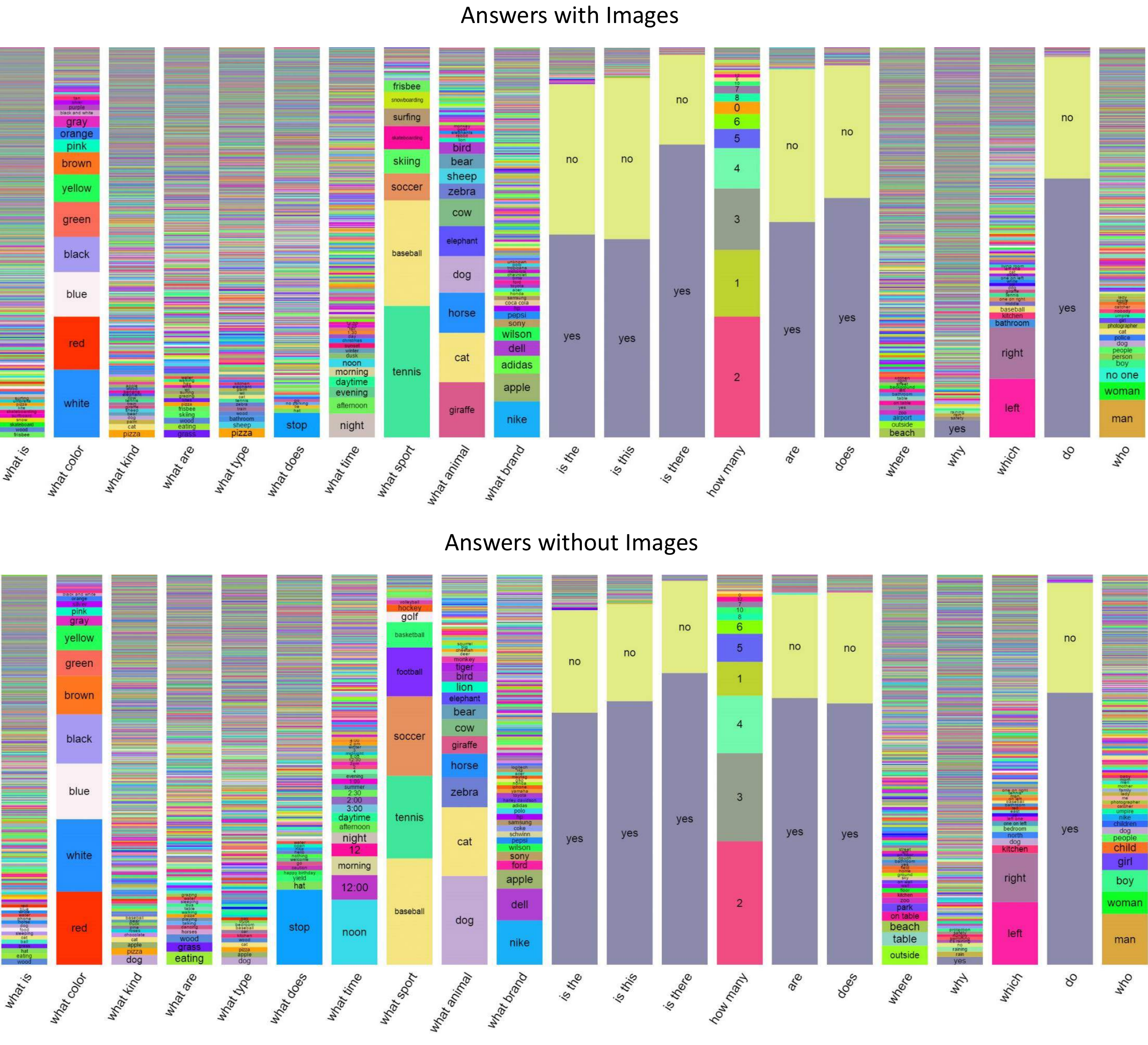}
\caption{Distribution of answers per question type for a random sample of 60K questions for real images when subjects provide answers when given the image (top) and when not given the image (bottom).}
\label{fig:AnsPerQues}
\end{figure*}

\subsection{Answers}

\textbf{Typical Answers.}
\figref{fig:AnsPerQues} (top) shows the distribution of answers for several question types.
We can see that a number of question types, such as ``Is the\ldots'', ``Are\ldots'', and ``Does\ldots'' are
typically answered using ``yes'' and ``no'' as answers.
Other questions such as ``What is\ldots'' and ``What type\ldots'' have a rich diversity
of responses. Other question types such as ``What color\ldots'' or ``Which\ldots'' have more specialized responses,
such as colors, or ``left'' and ``right''. 
See the appendix for a list of the most popular answers.

\textbf{Lengths.}
Most answers consist of a single word, with the distribution of answers containing one, two, or three words, respectively being $89.32\%$, $6.91\%$, and $2.74\%$ for real images and $90.51\%$, $5.89\%$, and $2.49\%$ for abstract scenes.
The brevity of answers is not surprising, since the questions tend to elicit specific
information from the images. This is in contrast with image captions that generically
describe the entire image and hence tend to be longer. The brevity of our answers makes
automatic evaluation feasible. While it may be tempting to believe the brevity of the answers
makes the problem easier, recall that they are human-provided open-ended answers to
open-ended questions. The questions typically require complex reasoning to arrive at these
deceptively simple answers (see \figref{fig:qualResults}).
There are currently 23,234 unique one-word answers in our dataset for real images and 3,770 for abstract scenes.

\textbf{`Yes/No' and `Number' Answers.}
Many questions are answered using either ``yes'' or ``no'' (or sometimes ``maybe'') -- 
$38.37\%$ and $40.66\%$ of the questions on real images and abstract scenes respectively. 
Among these `yes/no' questions, there is a bias towards 
``yes'' -- 
$58.83\%$ and $55.86\%$ of `yes/no' answers are ``yes'' for real images and abstract scenes. 
Question types such as ``How many\ldots'' are answered using numbers -- 
$12.31\%$ and $14.48\%$ of the questions on real images and abstract scenes are `number' questions. 
``2'' is the most popular answer among the `number' questions, making up 
$26.04\%$ of the `number' answers for real images and $39.85\%$ for abstract scenes. 

\begin{figure}[t]
\centering
\includegraphics[width=1\linewidth]{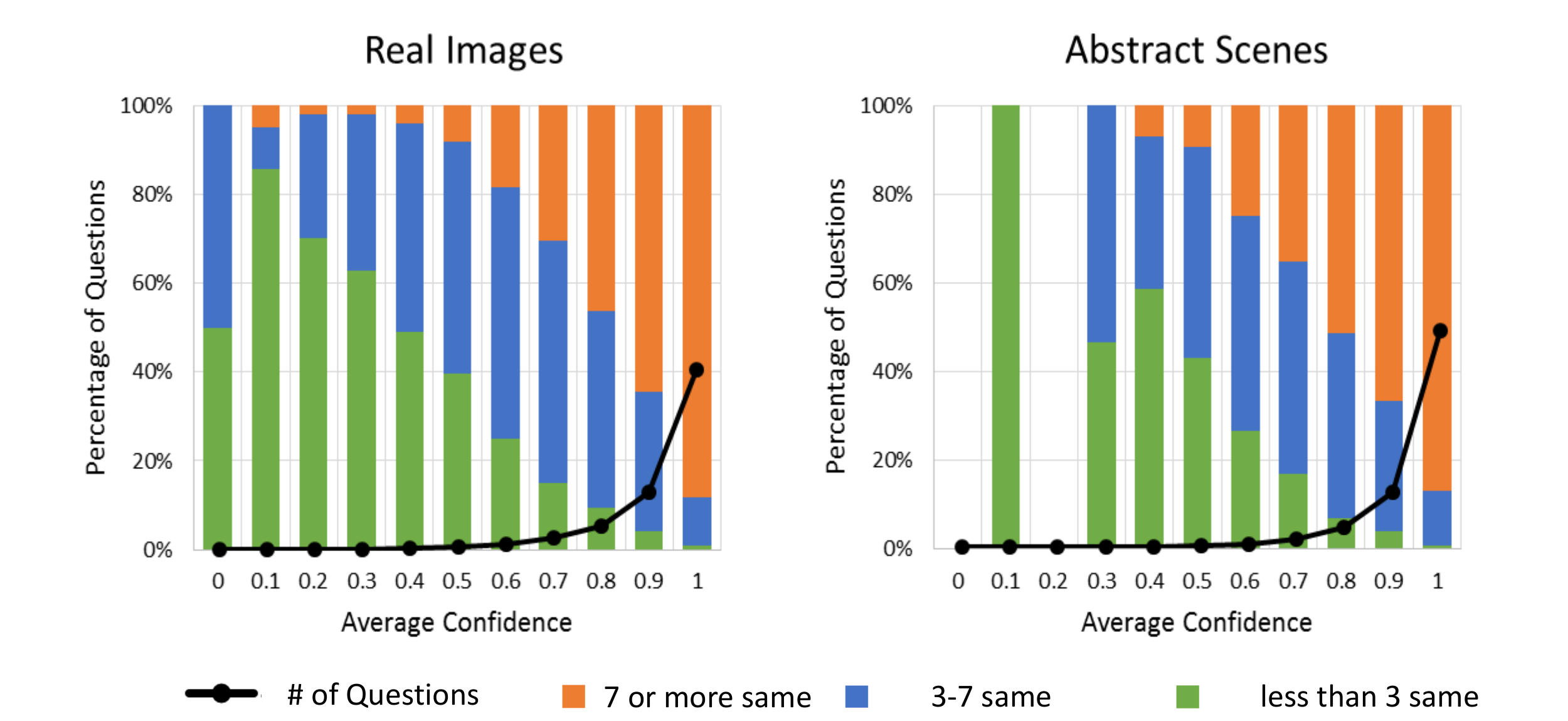}
\caption{Number of questions per average confidence score (0 = not confident, 1 = confident) for real images and abstract scenes (black lines). Percentage of questions where 7 or more answers are same, 3-7 are same, less than 3 are same (color bars). }
\label{fig:ConfScores}
\end{figure}

\textbf{Subject Confidence.}
When the subjects answered the questions, we asked
``Do you think you were able to answer the question correctly?''.
\figref{fig:ConfScores} shows the distribution of responses. A majority of the answers
were labeled as confident for both real images and abstract scenes. 

\textbf{Inter-human Agreement.}
Does the self-judgment of confidence correspond to the answer agreement between subjects?
\figref{fig:ConfScores} shows the percentage of questions in which 
(i) $7$ or more, 
(ii) $3-7$, or 
(iii) less than $3$ subjects agree on the answers given their average confidence score 
(0 = not confident, 1 = confident).
As expected, the agreement between subjects increases with confidence.
However, even if all of the subjects are confident the answers may still vary.
This is not surprising since some answers may vary, yet have very similar meaning, such as ``happy'' and ``joyful''.

As shown in \tableref{table:commonsense_acc} (Question + Image), there is significant inter-human
agreement in the answers for both real images ($83.30\%$) and abstract scenes ($87.49\%$). 
Note that on average each question has $2.70$ unique answers for real images and $2.39$ for abstract scenes. 
The agreement is significantly higher ($>95\%$) for \quotes{yes/no} questions and lower for other questions ($<76\%$), possibly due to the fact that we perform exact string matching and do not account for synonyms, plurality, \etc. Note that the automatic determination of synonyms is a difficult problem, since the level of answer granularity can vary across questions.

\subsection{Commonsense Knowledge}
\label{sec:cs}
\begin{figure*}[t]
 \includegraphics[width=\linewidth]{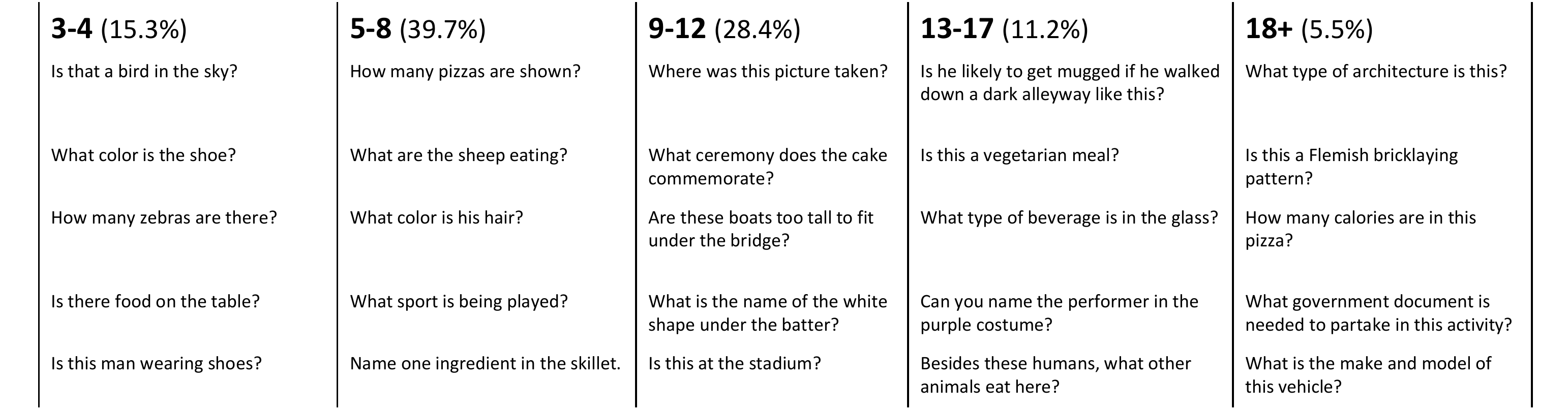}
 \centering
\caption{\small Example questions judged by Mturk workers to be answerable by different age groups. The percentage of questions falling into each age group is shown in parentheses.}
 \label{fig:age}
 \end{figure*}
 	
\textbf{Is the Image Necessary?}
Clearly, some questions can sometimes be
answered correctly using commonsense knowledge alone without the need for an image,
\eg, ``What is the color of the fire hydrant?''.
We explore this issue by asking three subjects to answer
the questions \emph{without seeing the image} (see the examples in blue in \figref{fig:qualResults}).
In \tableref{table:commonsense_acc} (Question), we show the percentage of questions for which
the correct answer is provided over all questions, ``yes/no'' questions, and the other questions that
are not ``yes/no''. For ``yes/no'' questions, the human subjects respond better than chance.
For other questions, humans are only correct about $21\%$ of the time. This demonstrates that
understanding the visual information is critical to VQA and that commonsense information alone is not sufficient.

To show the qualitative difference in answers provided with and without images,
we show the distribution of answers for various question types in \figref{fig:AnsPerQues} (bottom).
The distribution of colors, numbers, and even ``yes/no'' responses is surprisingly different for answers
with and without images.
 
\textbf{Which Questions Require Common Sense?}
In order to identify questions that require commonsense reasoning to answer, we conducted 
two AMT studies (on a subset 10K questions from the real images of VQA trainval) asking subjects --
\begin{compactenum} 
\item Whether or not they believed a question required commonsense to answer the question, and 
\item The youngest age group that they believe a person must be in order to be able to correctly answer the question -- 
toddler (3-4), 
younger child (5-8), 
older child (9-12), 
teenager (13-17), 
adult (18+).
\end{compactenum}
Each question was shown to 10 subjects. We found that 
for $47.43\%$ of questions 3 or more subjects voted `yes' to commonsense, 
($18.14\%$: 6 or more).  
In the `perceived human age required to answer question' study, we found the following distribution of responses: 
toddler: $15.3\%$,
younger child: $39.7\%$, 
older child: $28.4\%$, 
teenager: $11.2\%$, 
adult: $5.5\%$.
In Figure \ref{fig:age} we show several questions for which a majority of subjects picked the specified age range. Surprisingly the perceived age needed to answer the questions is fairly well distributed across the different age ranges. As expected the questions that were judged answerable by an adult (18+) generally need specialized knowledge, whereas those answerable by a toddler (3-4) are more generic.
 
We measure the degree of commonsense required to answer a question as the percentage of subjects (out of 10) who voted ``yes'' in our ``whether or not a question requires commonsense'' study.
A fine-grained breakdown of average age and average degree of common sense (on a scale of $0-100$) required to answer a question is shown in \tableref{tab:typeacc}. The average age and the average degree of commonsense across all questions is $8.92$ and $31.01\%$ respectively. 


It is important to distinguish between:
\begin{compactenum}
\item How old someone needs to be to be able to answer a question correctly,  and
\item How old people \emph{think} someone needs to be to be able to answer a question correctly. 
\end{compactenum}

Our age annotations capture the latter -- perceptions of MTurk workers in an uncontrolled environment. As such, the relative ordering of question types in \tableref{tab:typeacc} is more important than absolute age numbers.
The two rankings of questions in terms of common sense required according to the two studies 
were largely correlated (Pearson's rank correlation: 0.58). 

\begin{table}[t]
\setlength{\tabcolsep}{3.2pt}
{\small
\begin{center}
\begin{tabular}{@{}llcccc@{}}
\toprule
Dataset & Input & All & Yes/No & Number & Other \\
\midrule
    & Question & 40.81 & 67.60 & 25.77 & 21.22 \\
Real   & Question + Caption* & 57.47 & 78.97 & 39.68 & 44.41 \\
    & Question + Image & 83.30 & 95.77 & 83.39 & 72.67 \\
\midrule
 & Question & 43.27 & 66.65 & 28.52 & 23.66 \\
Abstract & Question + Caption* & 54.34 & 74.70 & 41.19 & 40.18 \\
 & Question + Image & 87.49 & 95.96 & 95.04 & 75.33 \\
\bottomrule
\end{tabular}
\end{center}
}
\caption {Test-standard accuracy of human subjects when asked to answer the 
question without seeing the image (Question), 
seeing just a caption of the image and not the image itself (Question + Caption), 
and seeing the image (Question + Image). 
Results are shown for all questions, ``yes/no'' \& ``number'' questions, and other questions 
that are neither answered ``yes/no'' nor number. 
All answers are free-form and not multiple-choice. 
*\hspace{1pt}These accuracies are evaluated on a subset of 3K train questions (1K images).}
\label{table:commonsense_acc}
\end{table}

\subsection{Captions \textbf{\vs} Questions}

Do generic image captions provide enough information to answer the questions?
\tableref{table:commonsense_acc} (Question + Caption) shows the percentage of questions answered
correctly when human subjects are given the question and a human-provided caption
describing the image, but not the image. As expected, the results are better than when humans are shown the questions alone.
However, the accuracies are significantly lower than when subjects are shown the actual image.
This demonstrates that in order to answer the questions correctly, deeper image understanding 
(beyond what image captions typically capture) is necessary. In fact, we find that the distributions of nouns, verbs, and adjectives mentioned in captions is statistically significantly different from those mentioned in our questions + answers (Kolmogorov-Smirnov test, $p<.001$) for both real images and abstract scenes. See the appendix for details. 

%% file: baselines.tex
\section{VQA Baselines and Methods}
\begin{figure*}[h]
\includegraphics[width=1\linewidth]{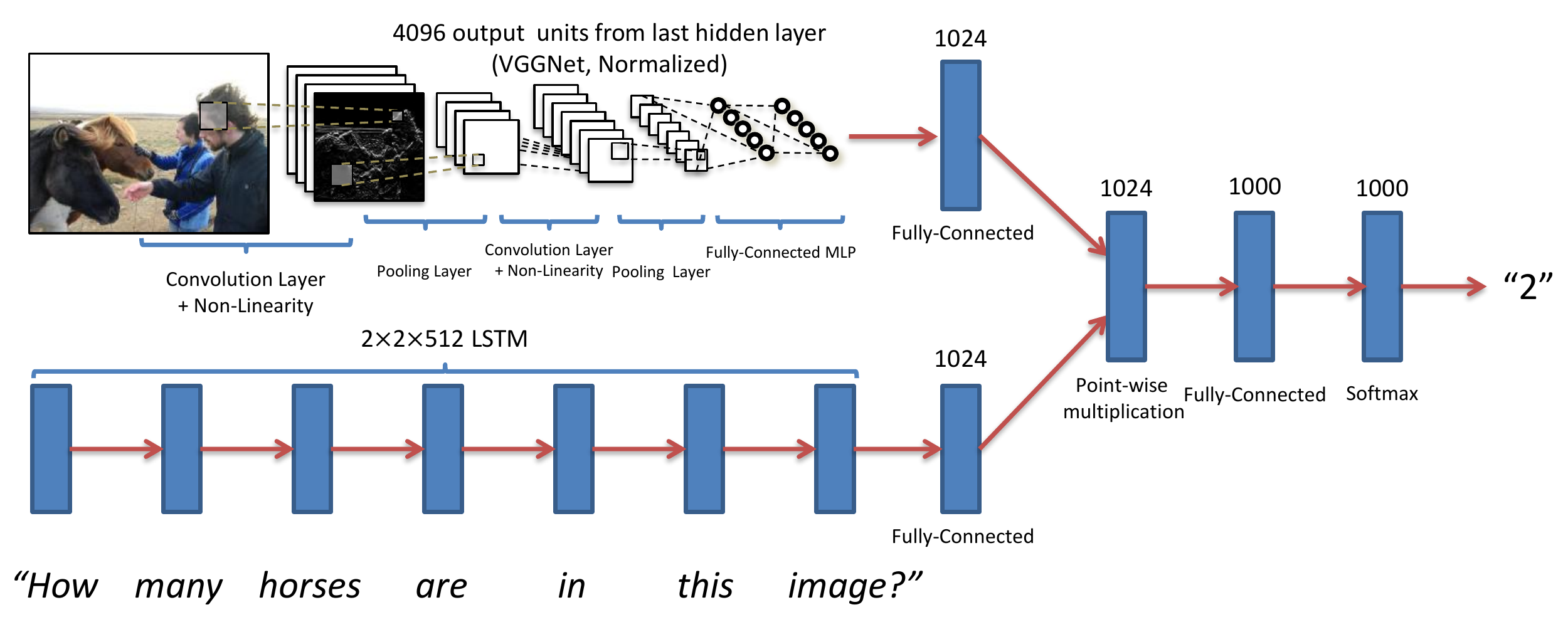}
\centering
\caption{Our best performing model (deeper LSTM Q + norm I). This model uses a two layer LSTM to encode the questions and the last hidden layer of VGGNet~\cite{Simonyan14c} to encode the images. The image features are then $\ell_2$ normalized. Both the question and image features are transformed to a common space and fused via element-wise multiplication, which is then passed through a fully connected layer followed by a softmax layer to obtain a distribution over answers.}
\label{fig:best_model}
\end{figure*}
In this section, we explore the difficulty of the VQA dataset for the MS COCO images using several baselines 
and novel methods. We train on VQA train+val. Unless stated otherwise, all human accuracies are on test-standard, machine accuracies are on test-dev, and results involving human captions (in gray font) are trained on train and tested on val (because captions are not available for test). 

\subsection{Baselines}
\label{sec:baselinesmain}
We implemented the following baselines:
\begin{enumerate}
\item \textbf{random:} We randomly choose an answer from the top 1K answers of the VQA train/val dataset.

\item \textbf{prior (``yes''):} We always select the most popular answer (``yes'') for both the open-ended and multiple-choice tasks. Note that ``yes'' is always one of the choices for the multiple-choice questions.

\item \textbf{per Q-type prior:} For the open-ended task, we pick the most popular answer per question type (see the appendix for details). For the multiple-choice task, we pick the answer (from the provided choices) that is most similar to the picked answer for the open-ended task using cosine similarity in Word2Vec\cite{word2vec} feature space.

\item \textbf{nearest neighbor:} Given a test image, question pair, we first find the $K$ nearest neighbor questions and associated images from the training set. See appendix for details on how neighbors are found. Next, for the open-ended task, we pick the most frequent ground truth answer from this set of nearest neighbor question, image pairs. Similar to the ``per Q-type prior'' baseline, for the multiple-choice task, we pick the answer (from the provided choices) that is most similar to the picked answer for the open-ended task using cosine similarity in \\ Word2Vec\cite{word2vec} feature space.
\end{enumerate}

\vspace{-5pt}
\subsection{Methods}
\label{sec:methods}
For our methods, we develop a 2-channel vision (image) + language (question) model that culminates with a softmax over $K$ possible outputs. We choose the top $K = 1000$ most frequent answers as possible outputs. This set of answers covers $82.67\%$ of the train+val answers. We describe the different components of our model below:

\textbf{Image Channel:} This channel provides an embedding for the image. We experiment with two embeddings -- 
\begin{compactenum}
\item \textbf{I:} The activations from the last hidden layer of VGGNet~\cite{Simonyan14c} are used as 4096-dim image embedding.
\item \textbf{norm I:} These are $\ell_2$ normalized activations from the last hidden layer of VGGNet~\cite{Simonyan14c}.
\end{compactenum}

\textbf{Question Channel:} This channel provides an embedding for the question. We experiment with three embeddings --
\begin{compactenum}
\item \textbf{Bag-of-Words Question (BoW Q)}: The top 1,000 words in the questions are used to create a bag-of-words representation. Since there is a strong correlation between the words that start a question and the answer (see \figref{fig:AnsPerQues}), we find the top 10 first, second, and third words of the questions and create a 30 dimensional bag-of-words representation. These features are concatenated to get a 1,030-dim embedding for the question.
\item \textbf{LSTM Q:} An LSTM with one hidden layer is used to obtain 1024-dim embedding for the question. The embedding obtained from the LSTM is a concatenation of last cell state and last hidden state representations (each being 512-dim) from the hidden layer of the LSTM. Each question word is encoded with 300-dim embedding by a fully-connected layer + tanh non-linearity which is then fed to the LSTM. The input vocabulary to the embedding layer consists of all the question words seen in the training dataset.
\item \textbf{deeper LSTM Q:} An LSTM with two hidden layers is used to obtain 2048-dim embedding for the question. The embedding obtained from the LSTM is a concatenation of last cell state and last hidden state representations (each being 512-dim) from each of the two hidden layers of the LSTM. Hence 2 (hidden layers) x 2 (cell state and hidden state) x 512 (dimensionality of each of the cell states, as well as hidden states) in \figref{fig:best_model}. This is followed by a fully-connected layer + tanh non-linearity to transform 2048-dim embedding to 1024-dim. The question words are encoded in the same way as in LSTM Q.   
\end{compactenum}

\textbf{Multi-Layer Perceptron (MLP):} The image and question embeddings 
are combined to obtain a single embedding. 
\begin{compactenum}
\item For \textbf{BoW Q + I} method, we simply concatenate the BoW Q and I embeddings. 
\item For \textbf{LSTM Q + I}, and \textbf{deeper LSTM Q + norm I} (\figref{fig:best_model}) methods, the image embedding is first transformed to 1024-dim by a fully-connected layer + tanh non-linearity to match the LSTM embedding of the question. The transformed image and LSTM embeddings (being in a common space) are then fused via element-wise multiplication. 
\end{compactenum}
This combined image + question embedding is then passed to an MLP -- a fully connected neural network classifier with 2 hidden layers and 1000 hidden units (dropout 0.5) in each layer with tanh non-linearity, followed by a softmax layer to obtain a distribution over $K$ answers. The entire model is learned end-to-end with a cross-entropy loss. VGGNet parameters are frozen to those learned for ImageNet classification and not fine-tuned in the image channel.   

We also experimented with providing captions as input to our model. Similar to \tableref{table:commonsense_acc}, we assume that a human-generated caption is given as input. We use a bag-of-words representation containing the 1,000 most popular words in the captions as the caption embedding (\textbf{Caption}). For \textbf{BoW Question + Caption (BoW Q + C)} method, we simply concatenate the BoW Q and C embeddings. 
For testing, we report the result on two different tasks: open-ended selects the answer with highest activation from all possible $K$ answers and multiple-choice picks the answer that has the highest activation from the potential answers. 

\subsection{Results}
\label{sec:results}

\begin{table}[t] \scriptsize
\setlength{\tabcolsep}{1.8pt}
\begin{center}
\begin{tabular}{@{} l  c  c  c  c  c  c c c@{}}
\toprule
& \multicolumn{4}{c}{Open-Ended} & \multicolumn{4}{c}{ Multiple-Choice} \\

\cmidrule[0.75pt](l){2-5}
\cmidrule[0.75pt](l){6-9}
 & All & Yes/No & Number & Other & All & Yes/No & Number & Other \\
\midrule
prior (``yes'') & 29.66 & 70.81 & 00.39 & 01.15 & 29.66 & 70.81 & 00.39 & 01.15 \\
per Q-type prior & 37.54 & 71.03 & 35.77 & 09.38 &  39.45 & 71.02 & 35.86 & 13.34 \\
nearest neighbor & 42.70 & 71.89 & 24.36 & 21.94 & 48.49 & 71.94 & 26.00 & 33.56 \\
BoW Q & \textcolor{black}{48.09} & \textcolor{black}{75.66} &  \textcolor{black}{36.70}& \textcolor{black}{27.14} 
& \textcolor{black}{53.68} & \textcolor{black}{75.71} & \textcolor{black}{37.05} & \textcolor{black}{38.64}\\
I & \textcolor{black}{28.13} & \textcolor{black}{64.01} & 00.42 & \textcolor{black}{03.77}  
&\textcolor{black}{30.53} & \textcolor{black}{69.87} & 00.45 & \textcolor{black}{03.76}\\
BoW Q + I & \textcolor{black}{52.64} & \textcolor{black}{75.55} & 33.67 & \textcolor{black}{37.37} 
& \textcolor{black}{58.97} & \textcolor{black}{75.59} & 34.35 & \textcolor{black}{50.33}\\
LSTM Q & \textcolor{black}{48.76} & \textcolor{black}{78.20} & 35.68 & \textcolor{black}{26.59} 
& \textcolor{black}{54.75} & \textcolor{black}{78.22} & 36.82 &\textcolor{black}{38.78}\\
LSTM Q + I & \textcolor{black}{53.74} & \textcolor{black}{78.94} & 35.24 & \textcolor{black}{36.42} 
& \textcolor{black}{57.17} & \textcolor{black}{78.95}& 35.80 & \textcolor{black}{43.41}\\
deeper LSTM Q & \textcolor{black}{50.39} & \textcolor{black}{78.41} & 34.68 & \textcolor{black}{30.03} 
& \textcolor{black}{55.88} & \textcolor{black}{78.45} & 35.91 &\textcolor{black}{41.13}\\
deeper LSTM Q + norm I & \textbf{57.75} & \textbf{80.50} & \textbf{36.77} & \textbf{43.08} 
& \textbf{62.70} & \textbf{80.52}& \textbf{38.22} & \textbf{53.01}\\
\midrule
\textcolor{gray}{Caption} & \textcolor{gray}{26.70} & \textcolor{gray}{65.50} & \textcolor{gray}{02.03} & \textcolor{gray}{03.86} 
& \textcolor{gray}{28.29} & \textcolor{gray}{69.79} & \textcolor{gray}{02.06} & \textcolor{gray}{03.82}\\
\textcolor{gray}{BoW Q + C} & \textcolor{gray}{54.70} & \textcolor{gray}{75.82} & \textcolor{gray}{40.12} & \textcolor{gray}{42.56} 
& \textcolor{gray}{59.85} & \textcolor{gray}{75.89} & \textcolor{gray}{41.16}  & \textcolor{gray}{52.53}\\
\bottomrule
\end{tabular}	
\caption{Accuracy of our methods for the open-ended and multiple-choice tasks on the VQA test-dev for real images. 
Q = Question, I = Image, C = Caption. (Caption and BoW Q + C results are on val). 
See text for details.
}
\label{tab:acc}
\end{center}
\end{table}
\tableref{tab:acc} shows the accuracy of our baselines and methods for both the open-ended and multiple-choice tasks on the VQA test-dev for real images.

As expected, the vision-alone model (I) that completely ignores the question performs rather poorly (open-ended: 28.13\% / multiple-choice: 30.53\%). In fact, on open-ended task, the vision-alone model (I) performs worse than the prior (``yes'') baseline, which ignores both the image \emph{and} question (responding to every question with a ``yes''). 

Interestingly, the language-alone methods (per Q-type prior, BoW Q, LSTM Q) that ignore the image perform surprisingly well, with BoW Q achieving 48.09\% on open-ended (53.68\% on multiple-choice) and LSTM Q achieving 48.76\% on open-ended (54.75\% on multiple-choice); both outperforming the nearest neighbor baseline (open-ended: 42.70\%, multiple-choice: 48.49\%). Our quantitative results and analyses suggest that this might be due to the language-model exploiting subtle statistical priors about the question types (e.g. ``What color is the banana?'' can be answered with ``yellow'' without looking at the image). For a detailed discussion of the subtle biases in the questions, please see \cite{yinyang}. 

The accuracy of our \textbf{best model} (deeper LSTM Q + norm I (\figref{fig:best_model}), selected using VQA test-dev accuracies) on VQA test-standard is 58.16\% (open-ended) / 63.09\% (multiple-choice). We can see that our model is able to significantly outperform both the vision-alone and language-alone baselines. As a general trend, results on multiple-choice are better than open-ended. All methods are significantly worse than human performance.

Our VQA demo is available on CloudCV \cite{cloudcv} -- \url{http://cloudcv.org/vqa}. This will be updated with newer models as we develop them.

To gain further insights into these results, we computed accuracies by question type in \tableref{tab:typeacc}. Interestingly, for question types that require more reasoning, such as ``Is the'' or ``How many'', the scene-level image features do not provide any additional information. However, for questions that can be answered using scene-level information, such as ``What sport,'' 
we do see an improvement. Similarly, for questions whose answer may be contained in a generic caption we see improvement, such as ``What animal''. For all question types, the results are worse than human accuracies.

\begin{figure*}[h]
\includegraphics[width=1\linewidth]{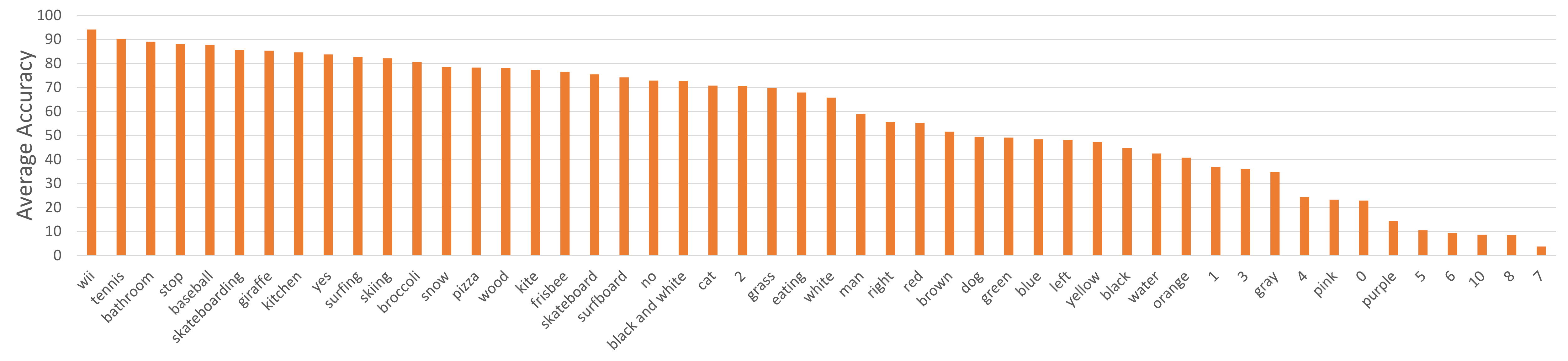}
\centering
\caption{$\Pr\text{(system is correct } | \text{ answer)}$ for 50 most frequent ground truth answers on the VQA validation set (plot is sorted by accuracy, not frequency). System refers to our best model (deeper LSTM Q + norm I).}
\label{fig:prob_1}
\end{figure*}

We also analyzed the accuracies of our best model (deeper LSTM Q + norm I) on a subset of questions with certain specific (ground truth) answers. In \figref{fig:prob_1}, we show the average accuracy of the model on questions with 50 most frequent ground truth answers on the VQA validation set (plot is sorted by accuracy, not frequency). We can see that the model performs well for answers that are common visual objects such as ``wii'', ``tennis'', ``bathroom'' while the performance is somewhat underwhelming for counts (\eg, ``2'', ``1'', ``3''), and particularly poor for higher counts (\eg, ``5'', ``6'', ``10'', ``8'', ``7''). 

\begin{figure*}[h]
\includegraphics[width=1\linewidth]{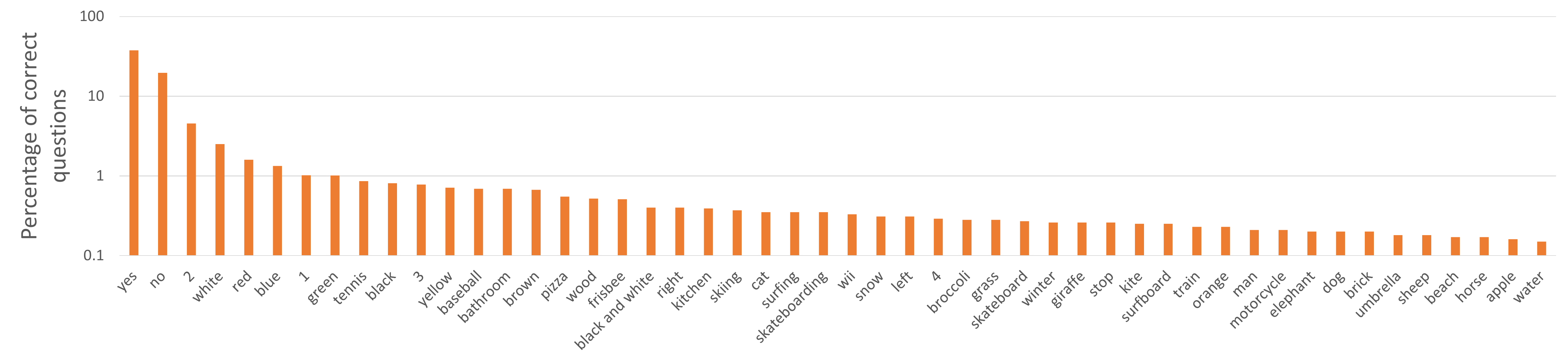}
\centering
\caption{$\Pr\text{(answer } | \text{ system is correct)}$ for 50 most frequently predicted answers on the VQA validation set (plot is sorted by prediction frequency, not accuracy). System refers to our best model (deeper LSTM Q + norm I).}
\label{fig:prob_2}
\end{figure*}

In \figref{fig:prob_2}, we show the distribution of 50 most frequently predicted answers when the system is correct on the VQA validation set (plot is sorted by prediction frequency, not accuracy). In this analysis, ``system is correct'' implies that it has VQA accuracy $1.0$ (see section \ref{sec:dataset} for accuracy metric). We can see that the frequent ground truth answers (\eg, ``yes'', ``no'', ``2'', ``white'', ``red'', ``blue'', ``1'', ``green'') are more frequently predicted than others when the model is correct. 



\begin{table}[h] \scriptsize
\setlength{\tabcolsep}{2pt}
\begin{center}
\begin{tabular}{@{} l  c  c  c  c c c c@{}  }
\toprule
& \multicolumn{5}{c}{Open-Ended}   & Human Age & Commonsense\\

\cmidrule[0.9pt](l){2-6}

\multicolumn{1}{c}{Question} & \multicolumn{3}{c}{K = 1000} & \multicolumn{2}{c}{Human} & To Be Able & To Be Able\\
\cmidrule[0.5pt](lr){2-4}
\cmidrule[0.5pt](l){5-6}

\multicolumn{1}{c}{Type}  & Q & Q + I & Q + C & Q & Q + I & To Answer & To Answer (\%)\\
\midrule
what is \textcolor{black}{(13.84)} & \textcolor{black}{23.57} & \textcolor{black}{34.28} & \textcolor{gray}{43.88} & \textcolor{black}{16.86} & \textcolor{black}{73.68} &\textcolor{black}{09.07} & 27.52\\
what color \textcolor{black}{(08.98)} & \textcolor{black}{33.37} & \textcolor{black}{43.53} & \textcolor{gray}{48.61} & \textcolor{black}{28.71} & \textcolor{black}{86.06} &\textcolor{black}{06.60} & 13.22\\
what kind \textcolor{black}{(02.49)} & \textcolor{black}{27.78} & \textcolor{black}{42.72} & \textcolor{gray}{43.88} & \textcolor{black}{19.10} & \textcolor{black}{70.11} &\textcolor{black}{10.55} & 40.34\\
what are \textcolor{black}{(02.32)} & \textcolor{black}{25.47} & \textcolor{black}{39.10} & \textcolor{gray}{47.27} & \textcolor{black}{17.72} & \textcolor{black}{69.49} &\textcolor{black}{09.03} & 28.72\\
what type \textcolor{black}{(01.78)} & \textcolor{black}{27.68} & \textcolor{black}{42.62} & \textcolor{gray}{44.32} & \textcolor{black}{19.53} & \textcolor{black}{70.65} &\textcolor{black}{11.04} & 38.92\\
is the \textcolor{black}{(10.16)} & \textcolor{black}{70.76} & \textcolor{black}{69.87} & \textcolor{gray}{70.50} & \textcolor{black}{65.24} & \textcolor{black}{95.67} &\textcolor{black}{08.51} & 30.30\\
is this \textcolor{black}{(08.26)} & \textcolor{black}{70.34} & \textcolor{black}{70.79} & \textcolor{gray}{71.54} & \textcolor{black}{63.35} & \textcolor{black}{95.43} &\textcolor{black}{10.13} & 45.32\\
how many \textcolor{black}{(10.28)} & \textcolor{black}{43.78} & \textcolor{black}{40.33} & \textcolor{gray}{47.52} & \textcolor{black}{30.45} & \textcolor{black}{86.32} &\textcolor{black}{07.67}& 15.93 \\
are \textcolor{black}{(07.57)} & \textcolor{black}{73.96} & \textcolor{black}{73.58} & \textcolor{gray}{72.43} & \textcolor{black}{67.10} & \textcolor{black}{95.24} &\textcolor{black}{08.65} & 30.63\\
does \textcolor{black}{(02.75)} & \textcolor{black}{76.81} & \textcolor{black}{75.81} & \textcolor{gray}{75.88} & \textcolor{black}{69.96} & \textcolor{black}{95.70} &\textcolor{black}{09.29} & 38.97\\
where \textcolor{black}{(02.90)} & \textcolor{black}{16.21} & \textcolor{black}{23.49} & \textcolor{gray}{29.47} & \textcolor{black}{11.09} & \textcolor{black}{43.56} &\textcolor{black}{09.54} & 36.51\\
is there \textcolor{black}{(03.60)} & \textcolor{black}{86.50} & \textcolor{black}{86.37} & \textcolor{gray}{85.88} & \textcolor{black}{72.48} & \textcolor{black}{96.43} &\textcolor{black}{08.25} & 19.88\\
why \textcolor{black}{(01.20)} & \textcolor{black}{16.24} & \textcolor{black}{13.94} & \textcolor{gray}{14.54} & \textcolor{black}{11.80} & \textcolor{black}{21.50} &\textcolor{black}{11.18} & 73.56\\
which \textcolor{black}{(01.21)} & \textcolor{black}{29.50} & \textcolor{black}{34.83} & \textcolor{gray}{40.84} & \textcolor{black}{25.64} & \textcolor{black}{67.44} &\textcolor{black}{09.27} & 30.00\\
do \textcolor{black}{(01.15)} & \textcolor{black}{77.73} & \textcolor{black}{79.31} & \textcolor{gray}{74.63} & \textcolor{black}{71.33} & \textcolor{black}{95.44} &\textcolor{black}{09.23} & 37.68\\
what does \textcolor{black}{(01.12)} & \textcolor{black}{19.58} & \textcolor{black}{20.00} & \textcolor{gray}{23.19} & \textcolor{black}{11.12} & \textcolor{black}{75.88} &\textcolor{black}{10.02} & 33.27\\
what time \textcolor{black}{(00.67)} & \textcolor{black}{8.35} & \textcolor{black}{14.00} & \textcolor{gray}{18.28} & \textcolor{black}{07.64} & \textcolor{black}{58.98} &\textcolor{black}{09.81} & 31.83\\
who \textcolor{black}{(00.77)} & \textcolor{black}{19.75} & \textcolor{black}{20.43} & \textcolor{gray}{27.28} & \textcolor{black}{14.69} & \textcolor{black}{56.93} &\textcolor{black}{09.49} & 43.82\\
what sport \textcolor{black}{(00.81)} & \textcolor{black}{37.96} & \textcolor{black}{81.12} & \textcolor{gray}{93.87} & \textcolor{black}{17.86} & \textcolor{black}{95.59} &\textcolor{black}{08.07} & 31.87\\
what animal \textcolor{black}{(00.53)} & \textcolor{black}{23.12} & \textcolor{black}{59.70} & \textcolor{gray}{71.02} & \textcolor{black}{17.67} & \textcolor{black}{92.51} &\textcolor{black}{06.75} & 18.04\\
what brand \textcolor{black}{(00.36)} & \textcolor{black}{40.13} & \textcolor{black}{36.84} & \textcolor{gray}{32.19} & \textcolor{black}{25.34} & \textcolor{black}{80.95} &\textcolor{black}{12.50} & 41.33\\
                                                                                                                                                             
\bottomrule
\end{tabular}
\caption{Open-ended test-dev results for different question types on real images (Q+C is reported on val).
Machine performance is reported using the bag-of-words representation for questions.
Questions types are determined by the one or two words that start the question. 
The percentage of questions for each type is shown in parentheses. 
Last and second last columns respectively show the average human age and average degree of commonsense required to answer the questions 
(as reported by AMT workers), respectively. 
See text for details.}
\vspace{-10pt}
\label{tab:typeacc}
\end{center}
\end{table}


Finally, evaluating our best model (deeper LSTM Q + norm I) on the validation questions for which we have age annotations (how old a human needs to be to answer the question correctly), we estimate that our model performs as well as a $4.74$ year old child! The average age required on the same set of questions is $8.98$. Evaluating the same model on the validation questions for which we have commonsense annotations (whether the question requires commonsense to answer it), we estimate that it has degree of commonsense of $17.35\%$. The average degree of commonsense required on same set of questions is $31.23\%$. Again, these estimates reflect the age and commonsense perceived by MTurk workers that would be required to answer the question. See the appendix for details. 

\begin{figure}[h]
\includegraphics[width=1\linewidth]{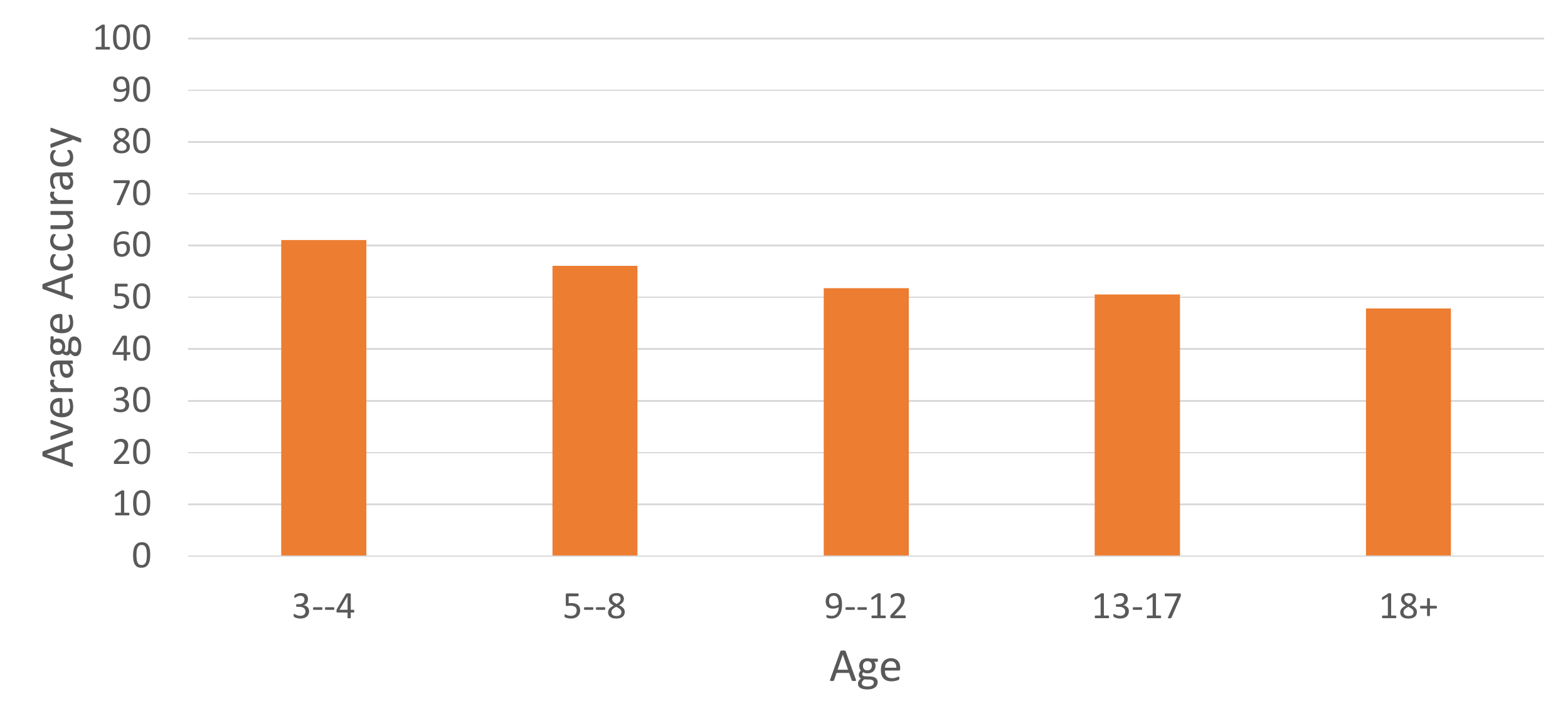}
\centering
\caption{$\Pr\text{(system is correct } | \text{ age of question)}$ on the VQA validation set. System refers to our best model (deeper LSTM Q + norm I).}
\label{fig:age_1}
\end{figure}

We further analyzed the performance of the model for different age groups on the validation questions for which we have age annotations. In \figref{fig:age_1}, we computed the average accuracy of the predictions made by the model for questions belonging to different age groups. Perhaps as expected, the accuracy of the model decreases as the age of the question increases (from $61.07\%$ at $3-4$ age group to $47.83\%$ at $18+$ age group). 

\begin{figure}[h]
\includegraphics[width=1\linewidth]{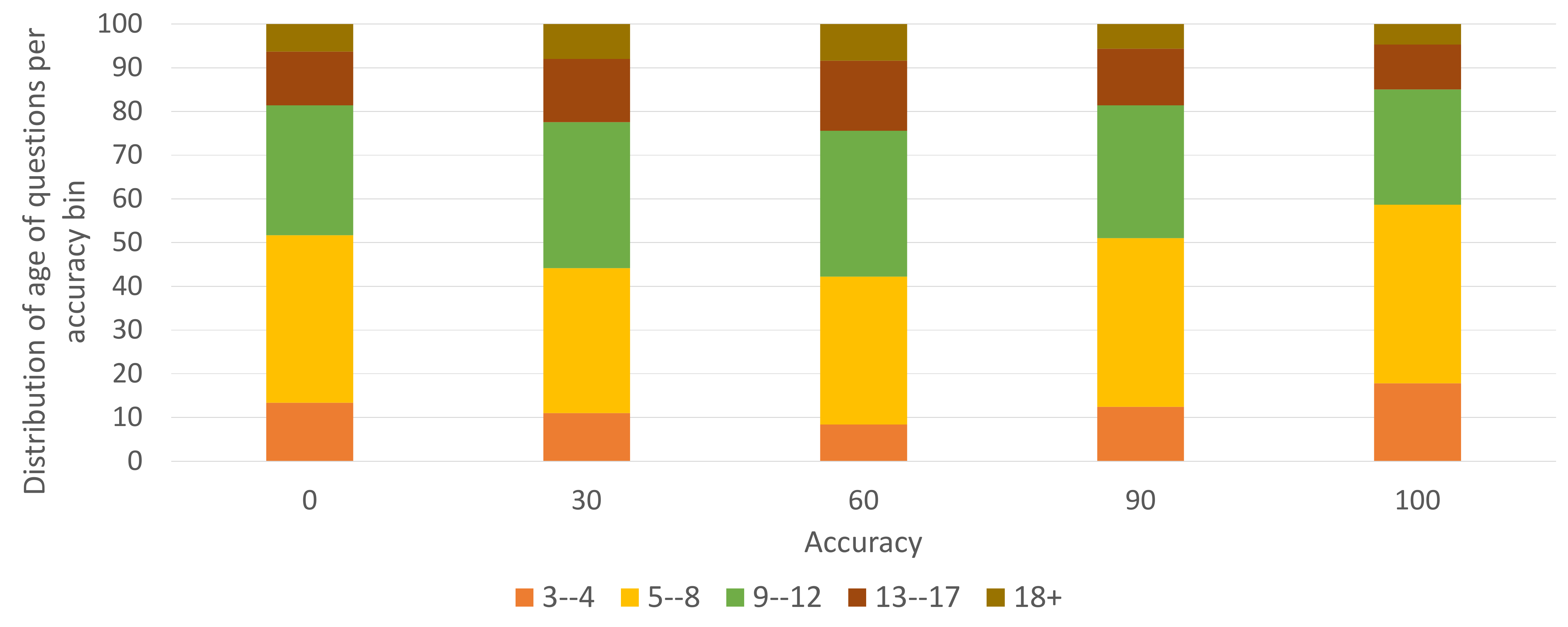}
\centering
\caption{$\Pr\text{(age of question } | \text{ system is correct)}$ on the VQA validation set. System refers to our best model (deeper LSTM Q + norm I).}
\label{fig:age_2}
\end{figure}

In \figref{fig:age_2}, we show the distribution of age of questions for different levels of accuracies achieved by our system on the validation questions for which we have age annotations. It is interesting to see that the relative proportions of different age groups is consistent across all accuracy bins with questions belonging to the age group 5-8 comprising the majority of the predictions which is expected because 5-8 is the most common age group in the dataset (see \figref{fig:age}).


\tableref{tab:abl_acc} shows the accuracy of different ablated versions of our best model (deeper LSTM Q + norm I) for both the open-ended and multiple-choice tasks on the VQA test-dev for real images. The different ablated versions are as follows --

\begin{compactenum}
\item \textbf{Without I Norm:} In this model, the activations from the last hidden layer of VGGNet~\cite{Simonyan14c} are not $\ell_2$-normalized. Comparing the accuracies in \tableref{tab:abl_acc} and \tableref{tab:acc}, we can see that $\ell_2$-normalization of image features boosts the performance by 0.16\% for open-ended task and by 0.24\% for multiple-choice task. 

\item \textbf{Concatenation:} In this model, the transformed image and LSTM embeddings are concatenated (instead of element-wise multiplied), resulting in doubling the number of parameters in the following fully-connected layer. Comparing the accuracies in \tableref{tab:abl_acc} and \tableref{tab:acc}, we can see that element-wise fusion performs better by 0.95\% for open-ended task and by 1.24\% for multiple-choice task.

\item \textbf{K = 500:} In this model, we use K = 500 most frequent answers as possible outputs. Comparing the accuracies in \tableref{tab:abl_acc} and \tableref{tab:acc}, we can see that K = 1000 performs better than K = 500 by 0.82\% for open-ended task and by 1.92\% for multiple-choice task.

\item \textbf{K = 2000:} In this model, we use K = 2000 most frequent answers as possible outputs. Comparing the accuracies in \tableref{tab:abl_acc} and \tableref{tab:acc}, we can see that K = 2000 performs better then K = 1000 by 0.40\% for open-ended task and by 1.16\% for multiple-choice task.

\item \textbf{Truncated Q Vocab $@$ 5:} In this model, the input vocabulary to the embedding layer (which encodes the question words) consists of only those question words which occur atleast 5 times in the training dataset, thus reducing the vocabulary size from 14770 (when all question words are used) to 5134 (65.24\% reduction). Remaining question words are replaced with UNK (unknown) tokens. Comparing the accuracies in \tableref{tab:abl_acc} and \tableref{tab:acc}, we can see that truncating the question vocabulary $@$ 5 performs better than using all questions words by 0.24\% for open-ended task and by 0.17\% for multiple-choice task.

\item \textbf{Truncated Q Vocab $@$ 11:} In this model, the input vocabulary to the embedding layer (which encodes the question words) consists of only those question words which occur atleast 11 times in the training dataset, thus reducing the vocabulary size from 14770 (when all question words are used) to 3561 (75.89\% reduction). Remaining question words are replaced with UNK (unknown) tokens. Comparing the accuracies in \tableref{tab:abl_acc} and \tableref{tab:acc}, we can see that truncating the question vocabulary $@$ 11 performs better than using all questions words by 0.06\% for open-ended task and by 0.02\% for multiple-choice task.

\item \textbf{Filtered Dataset:} We created a filtered version of the VQA train + val dataset in which we only keep the answers with subject confidence ``yes''. Also, we keep only those questions for which at least 50\% (5 out of 10) answers are annotated with subject confidence ``yes''. The resulting filtered dataset consists of 344600 questions, compared to 369861 questions in the original dataset, thus leading to only 6.83\% reduction in the size of the dataset. The filtered dataset has 8.77 answers per question on average. We did not filter the test set so that accuracies of the model trained on the filtered dataset can be compared with that of the model trained on the original dataset. The row ``Filtered Dataset'' in \tableref{tab:abl_acc} shows the performance of the deeper LSTM Q + norm I model when trained on the filtered dataset. Comparing these accuracies with the corresponding accuracies in \tableref{tab:acc}, we can see that the model trained on filtered version performs worse by 1.13\% for open-ended task and by 1.88\% for multiple-choice task.
\end{compactenum}

\begin{table}[t] \scriptsize
\setlength{\tabcolsep}{1.8pt}
\begin{center}
\begin{tabular}{@{} l  c  c  c  c  c  c c c@{}}
\toprule
& \multicolumn{4}{c}{Open-Ended} & \multicolumn{4}{c}{Multiple-Choice} \\

\cmidrule[0.75pt](l){2-5}
\cmidrule[0.75pt](l){6-9}
 & All & Yes/No & Number & Other & All & Yes/No & Number & Other \\
\midrule
Without I Norm & 57.59 & 80.41 & 36.63 & 42.84 & 62.46 & 80.43 & 38.10 & 52.62 \\
Concatenation & 56.80 & 78.49 & 35.08 & 43.19 & 61.46 & 78.52 & 36.43 & 52.54 \\
K = 500 & 56.93 & 80.61 & 36.24 & 41.39 & 60.78 & 80.64 & 37.44 & 49.10 \\
K = 2000 & 58.15 & 80.56 & 37.04 & 43.79 & 63.86 & 80.59 & 38.97 & 55.20 \\
Truncated Q Vocab $@$ 5 & 57.99 & 80.67 & 36.99 & 43.38 & 62.87 & 80.71 & 38.22 & 53.20\\
Truncated Q Vocab $@$ 11 & 57.81 & 80.42 & 36.97 & 43.22 & 62.72 & 80.45 & 38.30 & 53.09\\
Filtered Dataset & 56.62 & 80.19 & 37.48 & 40.95 & 60.82 & 80.19 & 37.48 & 49.57 \\
\bottomrule
\end{tabular}	
\caption{Accuracy of ablated versions of our best model (deeper LSTM Q + norm I) for the open-ended and multiple-choice tasks on the VQA test-dev for real images. 
Q = Question, I = Image. 
See text for details.
}
\label{tab:abl_acc}
\end{center}
\end{table}

%% file: challenge.tex
\section{VQA Challenge and Workshop}
\label{sec:challenge}

We have set up an evaluation server\footnote{\url{http://visualqa.org/challenge.html}} where results may be uploaded for the test set and it returns an accuracy breakdown. We are organizing an annual challenge and workshop to facilitate systematic progress in this area; the first instance of the workshop will be held at CVPR 2016\footnote{\url{http://www.visualqa.org/workshop.html}}. We suggest that papers reporting results on the VQA dataset --

\begin{compactenum}

\item Report test-standard accuracies, which can be calculated using either of the non-test-dev phases, i.e., ``test2015'' or ``Challenge test2015'' on the following links: [\href{https://www.codalab.org/competitions/6961}{oe-real} $|$ \href{https://www.codalab.org/competitions/6981}{oe-abstract} $|$ \href{https://www.codalab.org/competitions/6971}{mc-real} $|$ \href{https://www.codalab.org/competitions/6991}{mc-abstract}].

\item Compare their test-standard accuracies with those on the corresponding test2015 leaderboards [\href{http://www.visualqa.org/roe.html}{oe-real-leaderboard} $|$ \href{http://www.visualqa.org/aoe.html}{oe-abstract-leaderboard} $|$ \href{http://www.visualqa.org/rmc.html}{mc-real-leaderboard} $|$ \href{http://www.visualqa.org/amc.html}{mc-abstract-leaderboard}].

\end{compactenum}

\begin{figure*}[h]
\includegraphics[width=1\linewidth]{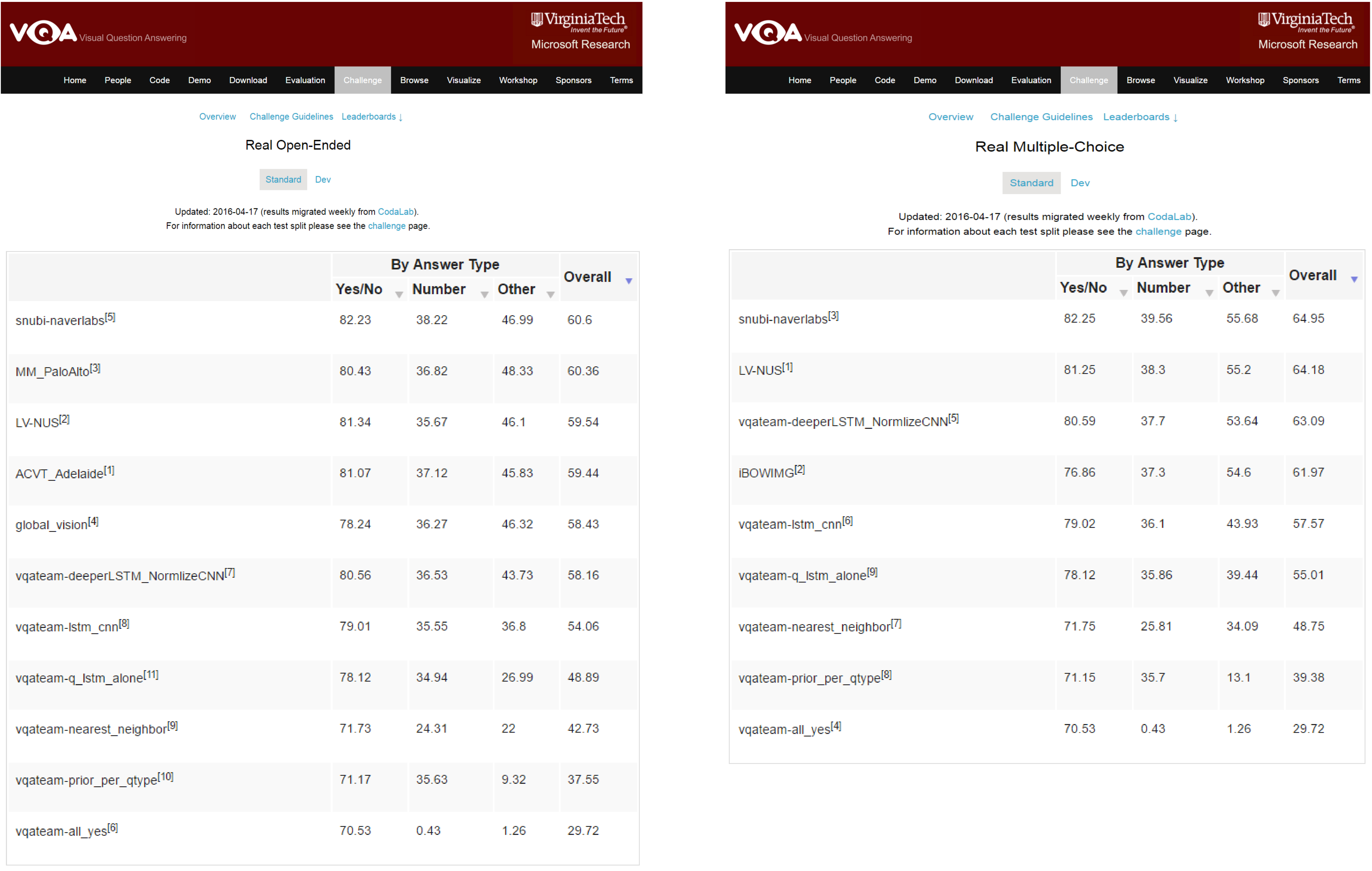}
\centering
\caption{Leaderboard showing test-standard accuracies for VQA Real Image Challenge (Open-Ended) on left and leaderboard showing test-standard accuracies for VQA Real Image Challenge (Multiple-Choice) on right (snapshot from \today).}
\label{fig:leaderboard-oe}
\end{figure*}

For more details, please see the challenge page\footnote{\url{http://visualqa.org/challenge.html}}. Screenshots of leaderboards for open-ended-real and multiple-choice-real are shown in \figref{fig:leaderboard-oe}.
We also compare the test-standard accuracies of our best model (deeper LSTM Q + norm I) for both open-ended and multiple-choice tasks (real images) with other entries (as of \today) on the corresponding leaderboards in \tableref{tab:comparison_leaderboard}.

\begin{table}[t] \scriptsize
\setlength{\tabcolsep}{1.8pt}
\begin{center}
\begin{tabular}{@{} l  c  c  c  c  c  c c c@{}}
\toprule
& \multicolumn{4}{c}{Open-Ended} & \multicolumn{4}{c}{ Multiple-Choice} \\

\cmidrule[0.75pt](l){2-5}
\cmidrule[0.75pt](l){6-9}
 & All & Yes/No & Number & Other & All & Yes/No & Number & Other \\
\midrule
snubi-naverlabs & 60.60 & 82.23 & 38.22 & 46.99 
& 64.95 & 82.25 & 39.56 & 55.68 \\
MM\_PaloAlto & 60.36 & 80.43 & 36.82 & 48.33 
& -- & -- & -- & -- \\
LV-NUS & 59.54 & 81.34 & 35.67 & 46.10 
& 64.18 & 81.25 & 38.30 & 55.20 \\
ACVT\_Adelaide & 59.44 & 81.07 & 37.12 & 45.83 
& -- & -- & -- & --\\
global\_vision & 58.43 & 78.24 & 36.27 & 46.32 
& -- & -- & -- & --\\
deeper LSTM Q + norm I & 58.16 & 80.56 & 36.53 & 43.73 
& 63.09 & 80.59 & 37.70 & 53.64\\
iBOWIMG	 & -- & -- & -- & --
& 61.97 & 76.86 & 37.30 & 54.60\\
\bottomrule
\end{tabular}	
\caption{Test-standard accuracy of our best model (deeper LSTM Q + norm I) compared to test-standard accuracies of other entries for the open-ended and multiple-choice tasks in the respective VQA Real Image Challenge leaderboards (as of \today).}
\label{tab:comparison_leaderboard}
\end{center}
\end{table}

%% file: discussion.tex
\section{Conclusion and Discussion}

In conclusion, we introduce the task of Visual Question Answering (VQA). Given an image and an open-ended, natural language question about the image, the task is to provide an accurate natural language answer. We provide a dataset containing over 250K images, 760K questions, and around 10M answers. We demonstrate the wide variety of questions and answers in our dataset, as well as the diverse set of AI capabilities in computer vision, natural language processing, and commonsense reasoning required to answer these questions accurately.

The questions we solicited from our human subjects were open-ended and not task-specific. For some application domains, it would be useful to collect task-specific questions. For instance, questions may be gathered from subjects who are visually impaired \cite{vizwiz}, or the questions could focused on one specific domain (say sports). 
Bigham \etal~\cite{vizwiz} created an application that allows the visually impaired to capture images and ask open-ended questions that are answered by human subjects. Interestingly, these questions can rarely be answered using generic captions. Training on task-specific datasets may help enable practical VQA applications.

We believe VQA has the distinctive advantage of pushing the frontiers on ``AI-complete'' problems, while being amenable to automatic evaluation. Given the recent progress in the community, we believe the time is ripe to take on such an endeavor. 

{
\textbf{Acknowledgements.}
We would like to acknowledge the countless hours of effort provided by the workers on Amazon Mechanical Turk. This work was supported in part by the The Paul G. Allen Family Foundation via an award to D.P., ICTAS at Virginia Tech via awards to D.B. and D.P., Google Faculty Research Awards to D.P. and D.B., the National Science Foundation CAREER award to D.B., the Army Research Office YIP Award to D.B., and a Office of Naval Research grant to D.B.
}

%% file: vqa_supplement_v2.tex
\section*{Appendix Overview}
In the appendix, we provide:
 \begin{enumerate}[I]
\setlength{\itemsep}{1pt}
  \setlength{\parskip}{0pt}
  \setlength{\parsep}{0pt}
 \item - Additional analysis comparing captions and Q\&A data
 \item - Qualitative visualizations for ``What is'' questions
 \item - Human accuracy on multiple-choice questions
 \item - \change{Details on VQA baselines}
  \item - \changenew{``Age'' and ``Commonsense'' of our model}
 \item - Details on the abstract scene dataset
 \item - User interfaces used to collect the dataset
 \item - List of the top answers in the dataset
 \item - Additional examples from the VQA dataset
 \end{enumerate}

%
\section*{Appendix I: Captions \vs Questions}
\label{sec:cap_vs_q}
Do questions and answers provide further information about the visual world beyond that captured by captions? One method for determining whether the information captured by questions \& answers is different from the information captured by captions is to measure some of the differences in the word distributions from the two datasets. We cast this comparison in terms of nouns, verbs, and adjectives by extracting all words from the caption data \change{(MS COCO captions for real images and captions collected by us for abstract scenes)} using the Stanford part-of-speech (POS)\footnote{Noun tags begin with NN, verb tags begin with VB, adjective tags begin with JJ, and prepositions are tagged as IN.} tagger~\cite{StanfordPOS}.  We normalize the word frequencies from captions, questions, and answers per image, and compare captions \vs questions and answers combined.  Using a Kolmogorov-Smirnov test to determine whether the underlying distributions of the two datasets differ, we find a significant difference for all three parts of speech (p $<$ .001) \change{for both real images and abstract scenes}. This helps motivate the VQA task as a way to learn information about visual scenes; although both captions and questions \& answers provide information about the visual world, they do it from different perspectives, with different underlying biases \cite{GordonVanDurme13}, and can function as complementary to one another.  

\begin{figure*}
\centering
\includegraphics[width=0.9\linewidth]{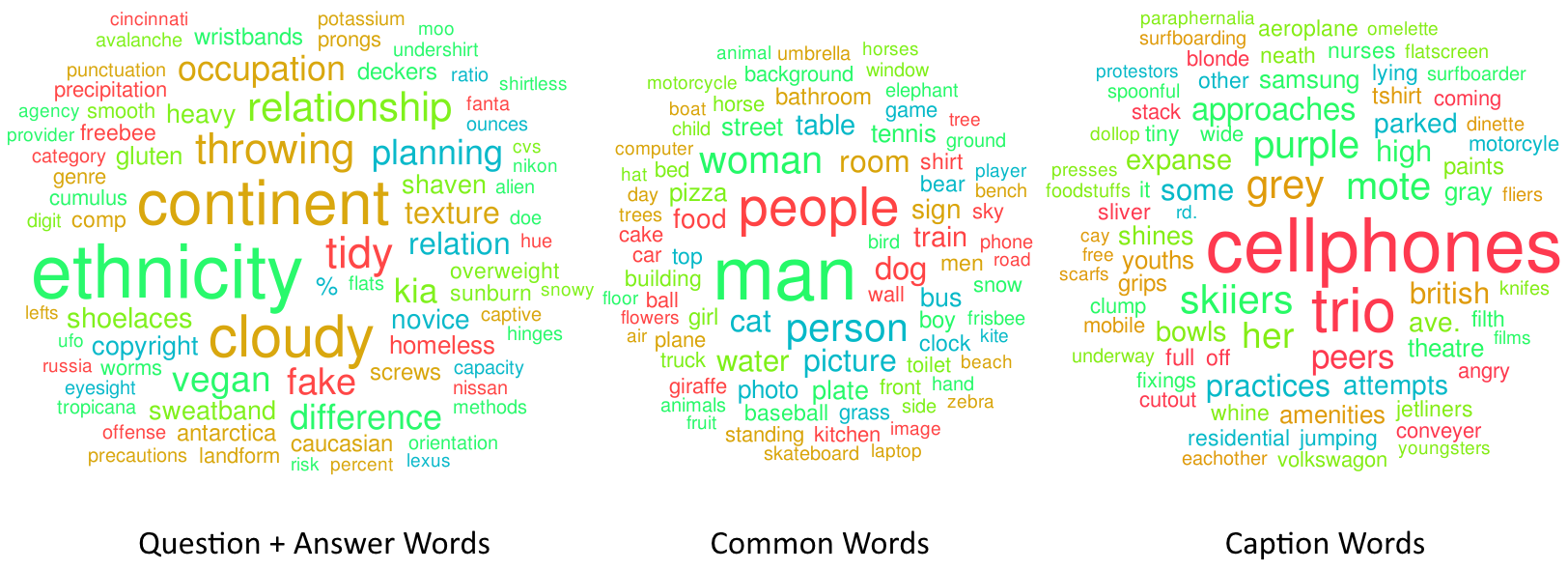}
\caption{Venn-style word clouds \cite{CoppersmithKelly14} for nouns with size indicating the normalized count \change{for real images}.}
\label{fig:noun_cloud_real}
\end{figure*}

\begin{figure*}
\centering
\includegraphics[width=0.9\linewidth]{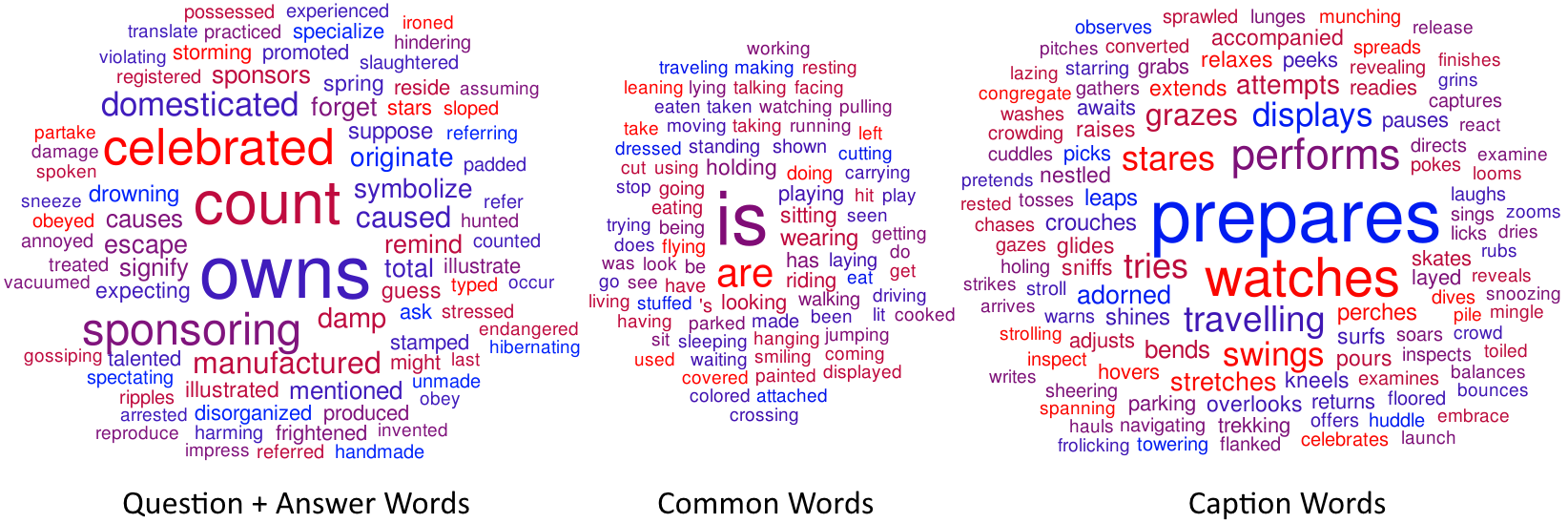}
\caption{Venn-style word clouds \cite{CoppersmithKelly14} for verbs with size indicating the normalized count \change{for real images}.}
\label{fig:verb_cloud_real}
\end{figure*}

\begin{figure*}
\centering
\includegraphics[width=0.9\linewidth]{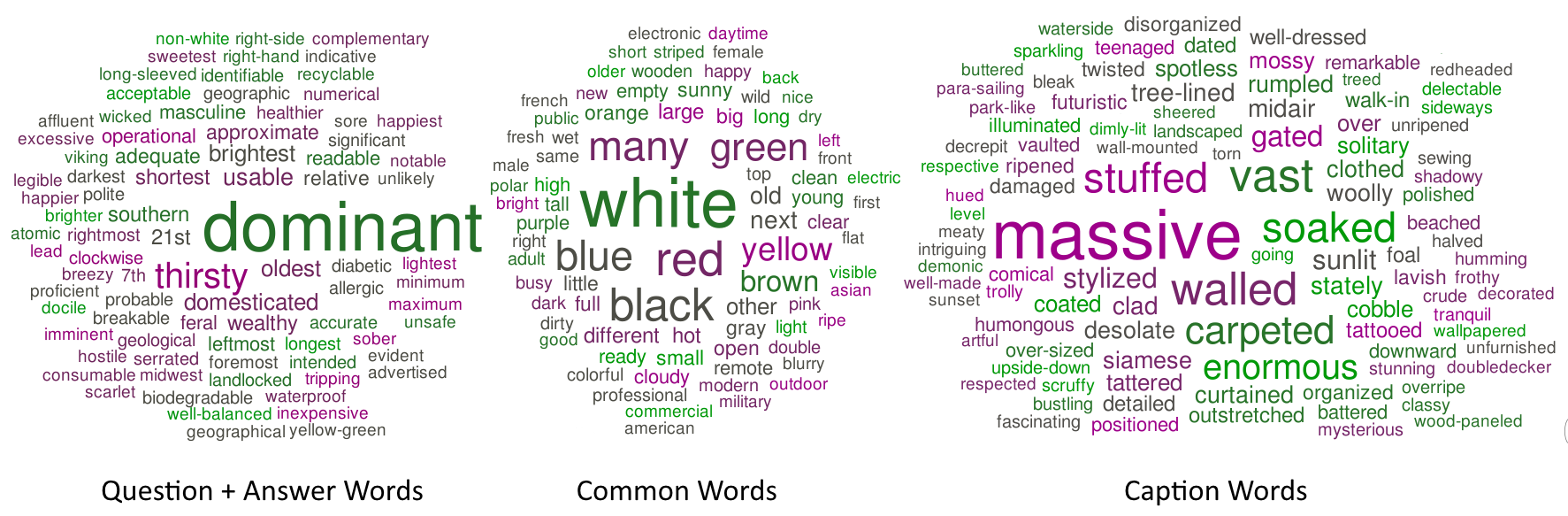}
\caption{Venn-style word clouds \cite{CoppersmithKelly14} for adjectives with size indicating the normalized count \change{for real images}.}
\label{fig:adj_cloud_real}
\end{figure*}

\begin{figure*}
\centering
\includegraphics[width=0.9\linewidth]{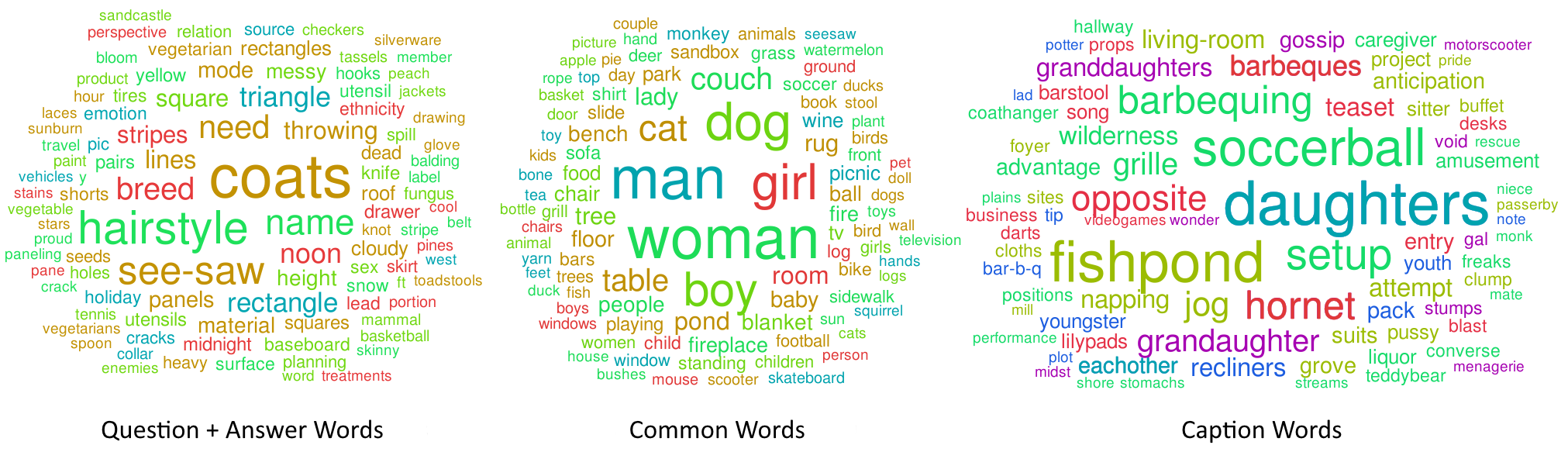}
\caption{Venn-style word clouds \cite{CoppersmithKelly14} for nouns with size indicating the normalized count \change{for abstract scenes}.}
\label{fig:noun_cloud_abs}
\end{figure*}

\begin{figure*}
\centering
\includegraphics[width=0.9\linewidth]{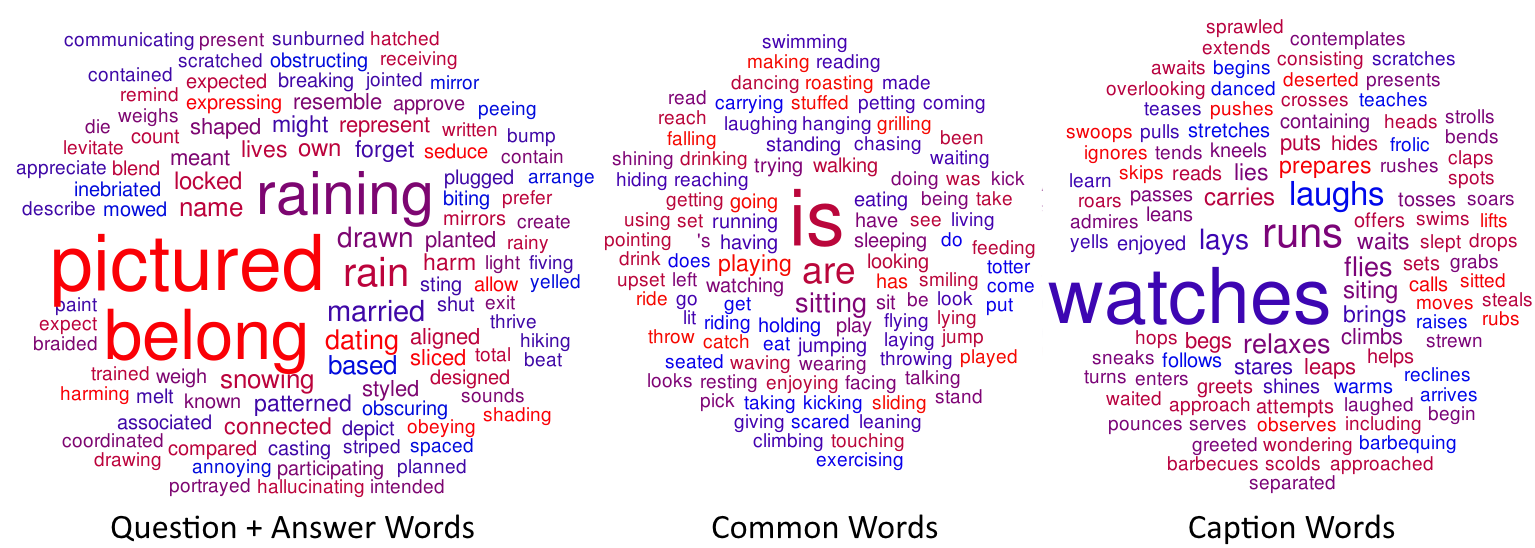}
\caption{Venn-style word clouds \cite{CoppersmithKelly14} for verbs with size indicating the normalized count \change{for abstract scenes}.}
\label{fig:verb_cloud_abs}
\end{figure*}

\begin{figure*}
\centering
\includegraphics[width=0.9\linewidth]{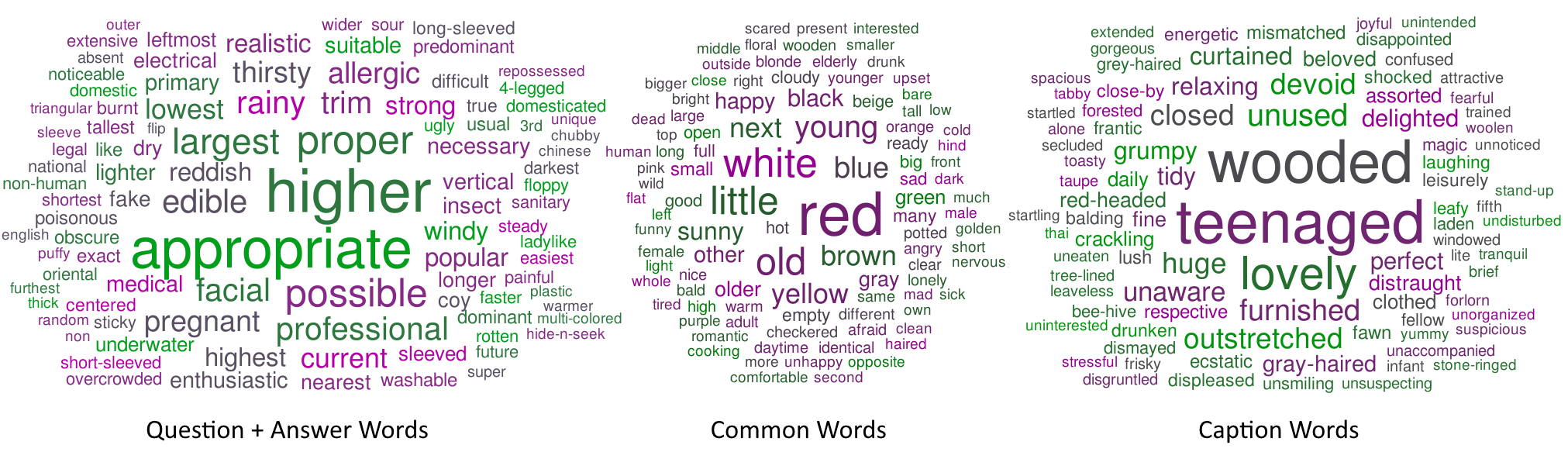}
\caption{Venn-style word clouds \cite{CoppersmithKelly14} for adjectives with size indicating the normalized count \change{for abstract scenes}.}
\label{fig:adj_cloud_abs}
\end{figure*}

We illustrate the similarities and differences between the word distributions in captions vs. questions \& answers as Venn-style word clouds \cite{CoppersmithKelly14} with size indicating the normalized count -- \figref{fig:noun_cloud_real} (nouns), \figref{fig:verb_cloud_real} (verbs), and \figref{fig:adj_cloud_real} (adjectives) \change{for real images and \figref{fig:noun_cloud_abs} (nouns), \figref{fig:verb_cloud_abs} (verbs), and \figref{fig:adj_cloud_abs} (adjectives) for abstract scenes}.\footnote{Visualization created using \url{http://worditout.com/}.}  The left side shows the top words in questions \& answers, the right the top words in captions, and the center the words common to both, with size indicating the harmonic mean of the counts.

We see that adjectives in captions capture some clearly visual properties discussed in previous work on vision to language \cite{MitchellEtAl13}, such as material and pattern, while the questions \& answers have more adjectives that capture what is usual (\eg, \change{``dominant'', ``approximate'', ``higher''}) and other kinds of commonsense properties (\eg, \change{``edible'', ``possible'', ``unsafe'', ``acceptable''}).  Interestingly, we see that question \& answer nouns capture information about \change{``ethnicity'' and ``hairstyle''}, while caption nouns capture information about \change{pluralized visible objects (\eg, ``cellphones'', ``daughters'') and groups (\eg, ``trio'', ``some''), among other differences.} ``Man'' and ``people'' are common in both captions and questions \& answers.

\begin{figure*}
\centering
\includegraphics[width=\linewidth]{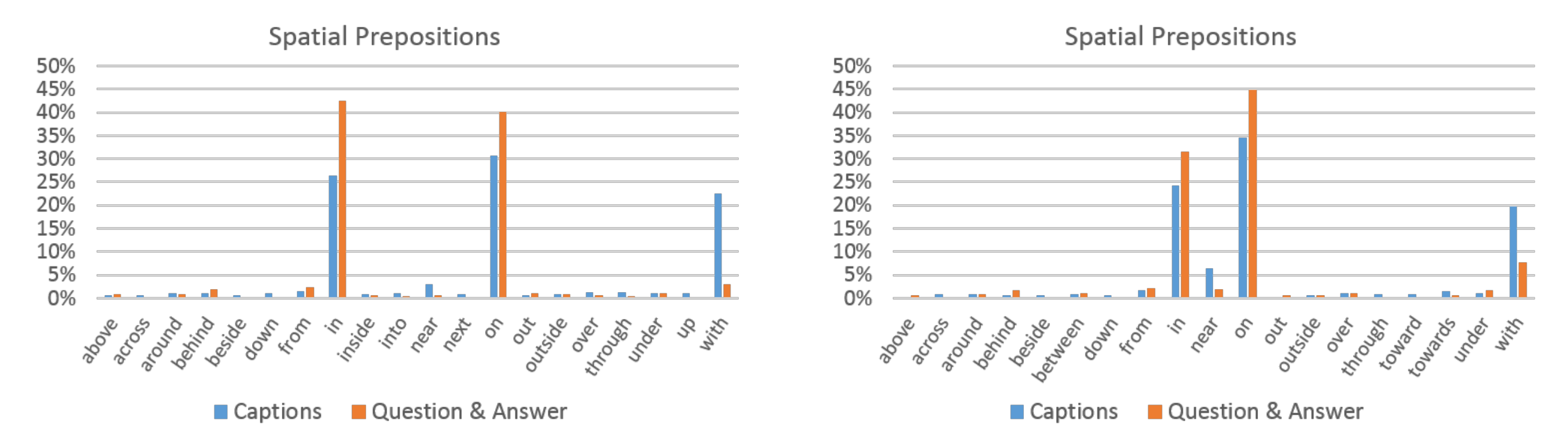}
\caption{Proportions of spatial prepositions in the captions and question \& answers for real images (left) and abstract scenes (right).}
\label{fig:spatial_prepositions}
\end{figure*}

One key piece to understanding the visual world is understanding spatial relationships, and so we additionally extract spatial prepositions and plot their proportions in the captions \vs~the questions \& answers data in \figref{fig:spatial_prepositions} (left) for real images and \figref{fig:spatial_prepositions} (right) for abstract scenes.  We see that questions \& answers have a higher proportion of specific spatial relations (\ie, ``in'', ``on'') compared to captions, which have a higher proportion of general spatial relations (\ie, ``with'', \change{``near''}).

\clearpage

\clearpage

\begin{figure*}
\centering
\includegraphics[width=0.95\linewidth]{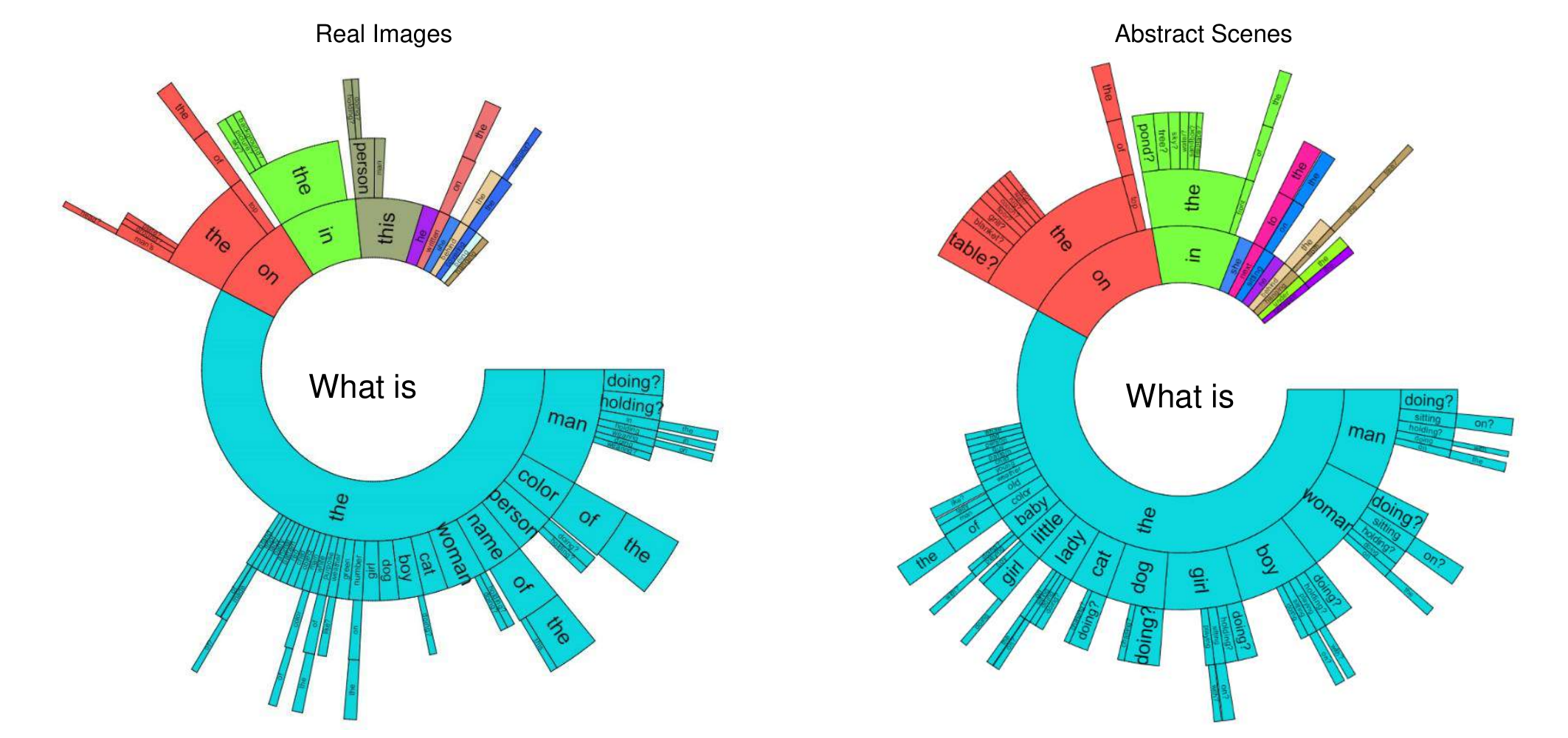}
\caption{\change{Distribution of questions starting with ``What is'' by their first five words for a random sample of 60K questions for real images (left) and all questions for abstract scenes (right). The ordering of the words starts towards the center and radiates outwards. The arc length is proportional to the number of questions containing the word. White areas are words with contributions too small to show.}}
\label{fig:WhatIsDistribution}
\end{figure*}
\begin{figure*}
\centering
\includegraphics[width=0.95\linewidth]{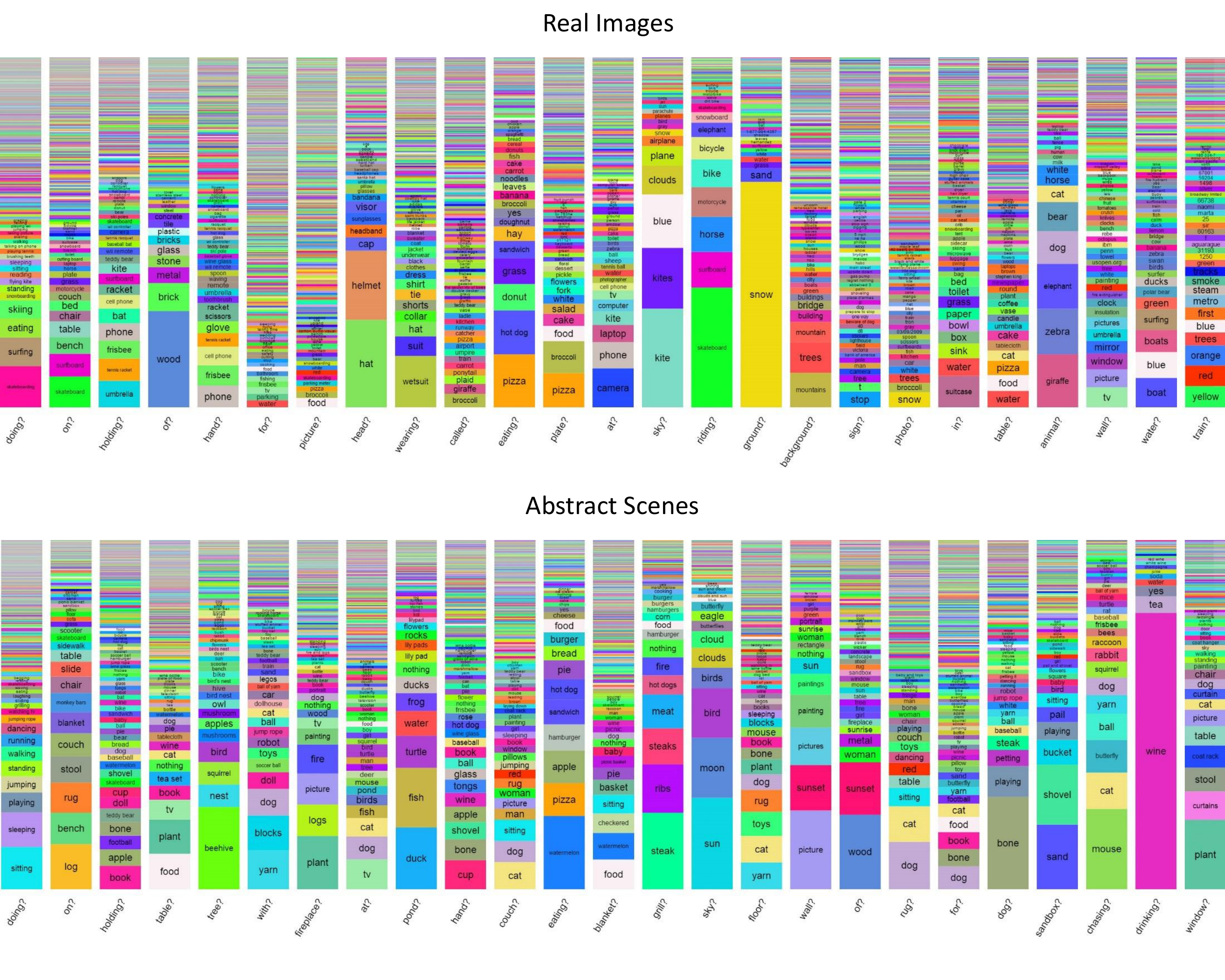}
\caption{\change{Distribution of answers for questions starting with ``What is'' for a random sample of 60K questions for real images (top) and all questions for abstract scenes (bottom). Each column corresponds to questions ending in different words, such as ``doing?'', ``on?'', \etc.}}
\label{fig:WhatIsAnswers}
\end{figure*}
\section*{Appendix II: ``What is'' Analysis}
\label{sec:what_is}

In \figref{fig:WhatIsDistribution}, we show the distribution of questions starting with ``What is'' by their first five words for \change{both real images and abstract scenes}. Note the diversity of objects referenced in the questions, as well as, the relations between objects, such as ``holding'' and \change{``sitting on''}. In \figref{fig:WhatIsAnswers}, we show the distribution of answers for ``What is'' questions ending in different words. For instance, questions ending in ``eating'' have answers such as ``pizza'',  \change{``watermelon'' and ``hot dog''}. Notice the diversity in answers for some questions, such as those that end with ``for?'' or \change{``picture?''}. Other questions result in intuitive responses, such as ``holding?'' and the response ``umbrella''.

\section*{Appendix III: Multiple-Choice Human Accuracy}
\label{sec:human_mc}

To compute human accuracy for multiple-choice questions, we collected \change{three} human answers per question on a random subset of 3,000 questions \change{for both real images and abstract scenes.} In \tableref{table:humanacc_mc}, we show the human accuracies for multiple choice questions. \tableref{table:humanacc_mc} also shows the inter-human agreement for open-ended answer task. In comparison to open-ended answer, the multiple-choice accuracies are 
\change{more or less same}
for ``yes/no'' questions and significantly better \change{($\approx15\%$ increase for real images and $\approx11\%$ increase for abstract scenes)} for ``other'' questions. Since ``other'' questions may be ambiguous, the increase in accuracy using multiple choice is not surprising. 

\begin{table}[t]
\setlength{\tabcolsep}{3pt}
{\small
\begin{center}
\begin{tabular}{@{}llcccc@{}}
\toprule
Dataset & Accuracy Metric & All & Yes/No & Number & Other \\
\midrule
 		& MC majority vote & \change{91.54} & \change{97.40} & \change{86.97} & \change{87.91} \\
  Real  & MC average & \change{88.53} & \change{94.40} & \change{84.99} & \change{84.64} \\
     	& \changenew{Open-Ended} & \changenew{80.62} & \changenew{94.78} & \changenew{78.46} & \changenew{69.69} \\
    \midrule
    	& MC majority vote & \change{93.57} & \change{97.78} & \change{96.71} & \change{88.73} \\ Abstract & MC average & \change{90.40} & \change{94.59} & \change{94.36} & \change{85.32} \\ 
    	& \changenew{Open-Ended} & \changenew{85.66} & \changenew{95.32} & \changenew{94.17} & \changenew{74.12} \\

    
\bottomrule
\end{tabular}
\end{center}
}
\caption {\changenew{For each of the two datasets, real and abstract, first two rows are the human accuracies for multiple-choice questions when subjects were shown both the image and the question. Majority vote means we consider the answer picked by majority of the three subjects to be the predicted answer by humans and compute accuracy of that answer for each question. Average means we compute the accuracy of each of the answers picked by the subjects and record their average for each question. The last row is the inter-human agreement for open-ended answers task when subjects were shown both the image and the question. All accuracies are evaluated on a random subset of 3000 questions.}}
\label{table:humanacc_mc}
\end{table}

\section*{Appendix IV: Details on VQA baselines}
\label{sec:baselines}
\textbf{``per Q-type prior'' baseline.} We decide on different question types based on first few words of questions in the real images training set and ensure that each question type has at least 30 questions in the training dataset. The most popular answer for each question type is also computed on real images training set. 

\textbf{``nearest neighbor'' baseline.} For every question in the VQA test-standard set, we find its $k$ nearest neighbor questions in the training set using cosine similarity in Skip-Thought \cite{kiros2015skip} feature space. We also experimented with bag of words and Word2Vec \cite{word2vec} feature spaces but we obtained the best performance with Skip-Thought. In this set of $k$ questions and their associated images, we find the image which is most similar to the query image using cosine similarity in fc7 feature space. We use the fc7 features from the caffenet model in BVLC Caffe \cite{jia2014caffe}. The most common ground truth answer of this most similar image and question pair is the predicted answer for the query image and question pair. We pick $k = 4$ on the test-dev set.

\section*{Appendix V: ``Age'' and ``Commonsense'' of our model}
\label{sec:model_age}
We estimate the age and degree of commonsense of our \textbf{best model} (deeper LSTM Q + norm I), selected using VQA test-dev accuracies). To estimate the age, we compute a weighted average of the average age per question, weighted by the accuracy of the model's predicted answer for that question, on the subset of questions in the VQA validation set for which we have age annotations (how old a human needs to be to answer the question correctly). To estimate the degree of commonsense, we compute a weighted average of the average degree of commonsense per question, weighted by the accuracy of the model's predicted answer for that question, on the subset of questions in the VQA validation set for which we have commonsense annotations (whether the question requires commonsense to answer it).


\section*{Appendix VI: Abstract Scenes Dataset}
\label{sec:abstract_scenes}
\begin{figure*}[h]
\centering
\includegraphics[width=1\linewidth]{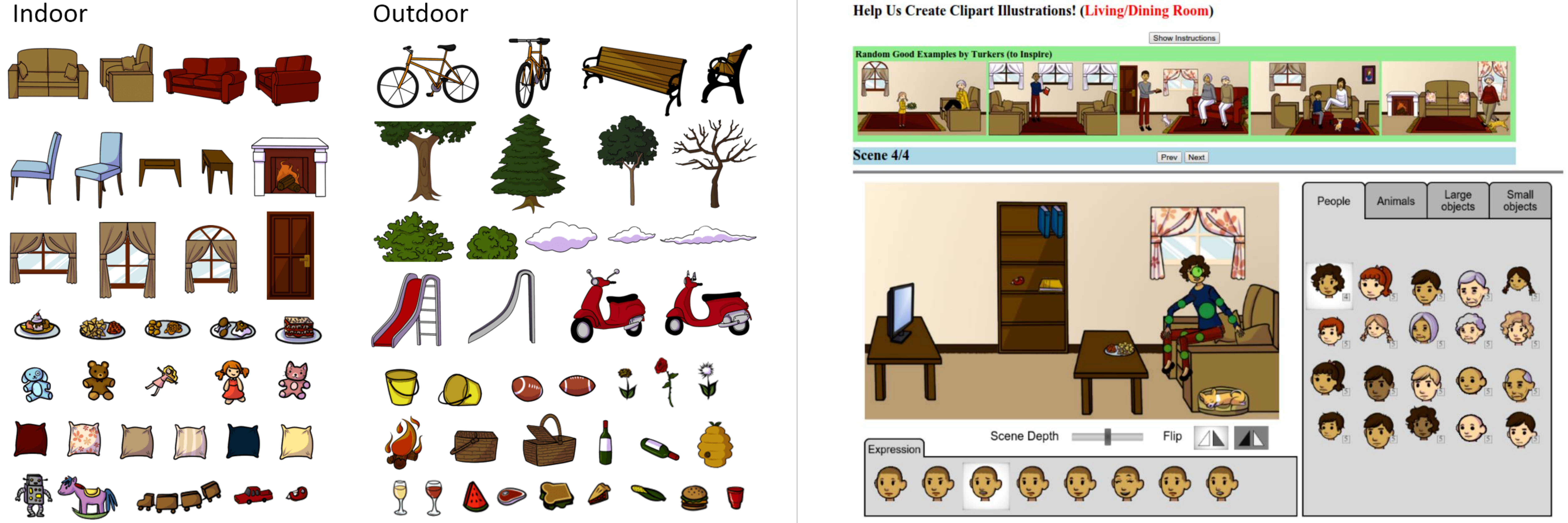}
\caption{Left: A small subset of the objects present in the abstract scene dataset. Right: The AMT interface for collecting abstract scenes.
The light green circles indicate where users can select to manipulate a person's pose. Different objects may be added to the scene using the folders to the right.}
\label{fig:clipart_details}
\vspace{10pt}
\end{figure*}

In \figref{fig:clipart_details} (left), we show a subset of the objects that are present in the abstract scenes dataset. For more examples of the scenes generated, please see \figref{fig:abstract_more_examples}. The user interface used to create the scenes is shown in \figref{fig:clipart_details} (right). Subjects used a drag-and-drop interface to create the scenes. Each object could be flipped horizontally and scaled. The scale of the object determined the rendering order of the objects. Many objects have different attributes corresponding to different poses or types. Most animals have five different discrete poses. Humans have eight discrete expressions and their poses may be continuously adjusted using a ``paperdoll'' model \cite{Antol2014}.

\section*{Appendix VII: User Interfaces}
\label{sec:uis}
\begin{figure*}[h]
\centering
\includegraphics[width=1\linewidth]{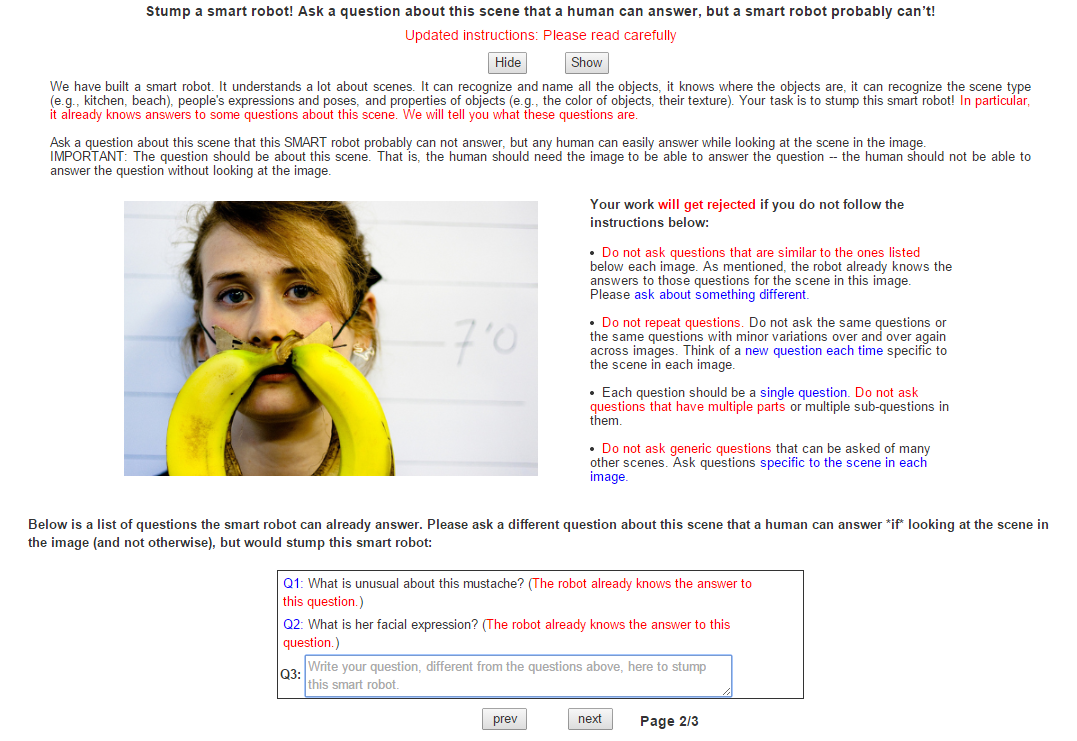}
\caption{Our AMT interface for collecting the third question for an image, when subjects were shown previous questions that were collected and were asked to ask a question different from previous questions.}
\label{fig:qstage03}
\end{figure*}

In \figref{fig:qstage03}, we show the AMT interface that we used to collect questions for images. Note that we tell the workers that the robot already knows the answer to the previously asked question(s), inspiring them to ask different kinds of questions, thereby increasing the diversity of our dataset.

\begin{figure*}[h]
\centering
\includegraphics[width=1\linewidth]{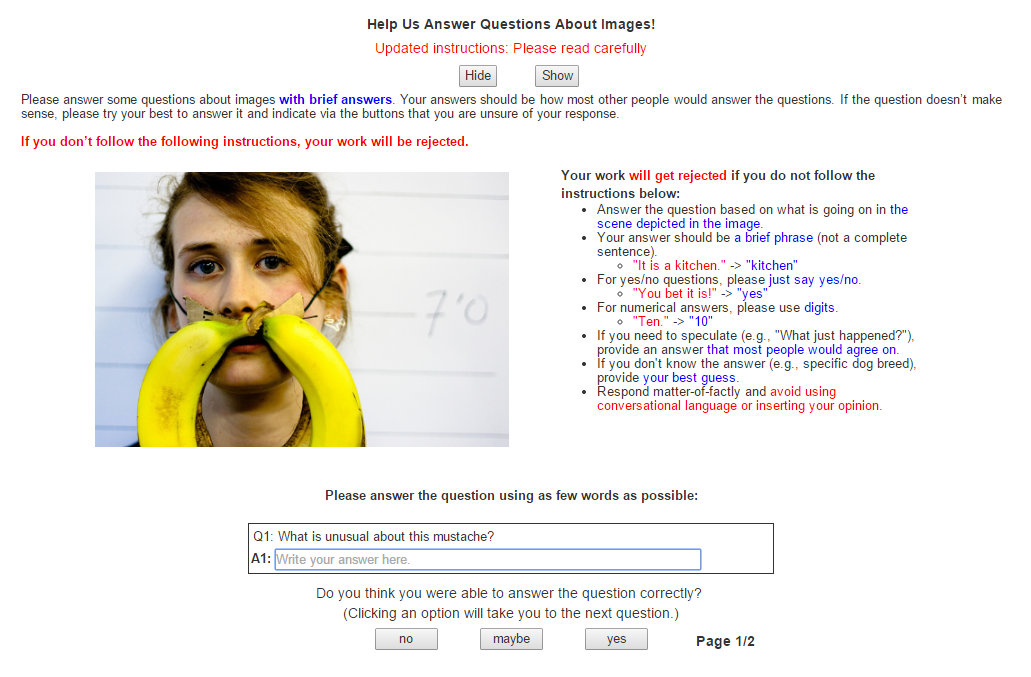}
\caption{The AMT interface used to collect answers to a question when subjects were shown the image while answering the question.}
\label{fig:answimage}
\end{figure*}

\begin{figure*}[h]
\centering
\includegraphics[width=1\linewidth]{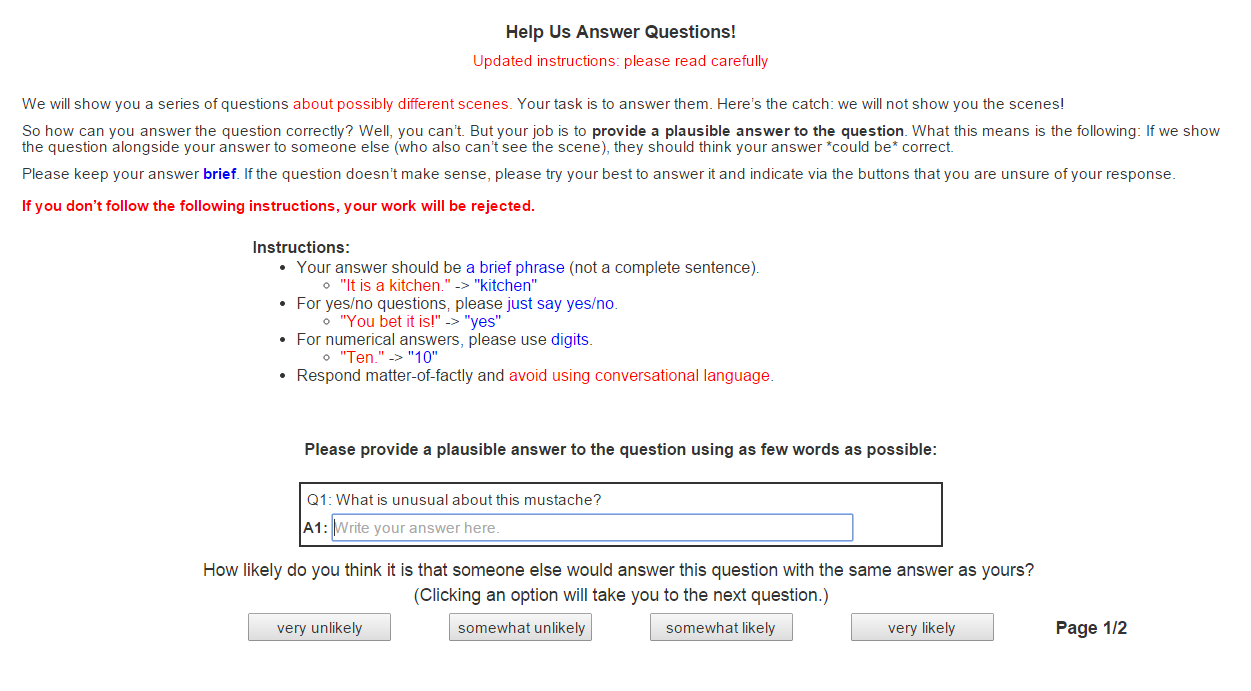}
\caption{The AMT interface used to collect answers to a question when subjects were not shown the image while answering the question using only commonsense to collect the plausible, but incorrect, multiple-choice answers.}
\label{fig:answoimage}
\end{figure*}

\figref{fig:answimage} shows the AMT interface used for collecting answers to the previously collected questions when subjects were shown the corresponding images. \figref{fig:answoimage} shows the interface that was used to collect answers to questions when subjects were not shown the corresponding image (\ie, to help in gathering incorrect, but plausible, answers for the multiple-choice task and to assess how accurately the questions can be answered using common sense knowledge alone).

\clearpage

\section*{Appendix VIII: Answer Distribution}
\label{sec:top_ans}
\vspace*{-8cm}
The top $250$ answers in our \change{real images} dataset along with their counts and percentage counts are given below. The answers have been presented in different colors to show the different Part-of-Speech (POS) tagging of the answers with the following color code: {\textcolor{magenta}{yes/no}}, {\textcolor{green}{noun}}, {\textcolor{blue}{verb}}, {\textcolor{yellow}{adjective}}, {\textcolor{cyan}{adverb}}, and {\textcolor{red}{numeral}}.

{\textcolor{magenta}{``yes''}} (566613, 22.82\%), {\textcolor{magenta}{``no''}} (381307, 15.35\%), {\textcolor{red}{``2''}} (80031, 3.22\%), {\textcolor{red}{``1''}} (46537, 1.87\%), {\textcolor{yellow}{``white''}} (41753, 1.68\%), {\textcolor{red}{``3''}} (41334, 1.66\%), {\textcolor{yellow}{``red''}} (33834, 1.36\%), {\textcolor{yellow}{``blue''}} (28881, 1.16\%), {\textcolor{red}{``4''}} (27174, 1.09\%), {\textcolor{yellow}{``green''}} (22453, 0.9\%), {\textcolor{yellow}{``black''}} (21852, 0.88\%), {\textcolor{yellow}{``yellow''}} (17312, 0.7\%), {\textcolor{yellow}{``brown''}} (14488, 0.58\%), {\textcolor{red}{``5''}} (14373, 0.58\%), {\textcolor{green}{``tennis''}} (10941, 0.44\%),{\textcolor{green}{``baseball''}} (10299, 0.41\%), {\textcolor{red}{``6''}} (10103, 0.41\%), {\textcolor{green}{``orange''}} (9136, 0.37\%), {\textcolor{red}{``0''}} (8812, 0.35\%), {\textcolor{green}{``bathroom''}} (8473, 0.34\%), {\textcolor{green}{``wood''}} (8219, 0.33\%), {\textcolor{cyan}{``right''}} (8209, 0.33\%), {\textcolor{cyan}{``left''}} (8058, 0.32\%), {\textcolor{green}{``frisbee''}} (7671, 0.31\%), {\textcolor{yellow}{``pink''}} (7519, 0.3\%), {\textcolor{yellow}{``gray''}} (7385, 0.3\%), {\textcolor{green}{``pizza''}} (6892, 0.28\%), {\textcolor{red}{``7''}} (6005, 0.24\%), {\textcolor{green}{``kitchen''}} (5926, 0.24\%), {\textcolor{red}{``8''}} (5592, 0.23\%), {\textcolor{green}{``cat''}} (5514, 0.22\%), {\textcolor{green}{``skiing''}} (5189, 0.21\%), {\textcolor{blue}{``skateboarding''}} (5122, 0.21\%), {\textcolor{green}{``dog''}} (5092, 0.21\%), {\textcolor{green}{``snow''}} (4867, 0.2\%), {\textcolor{yellow}{``black and white''}} (4852, 0.2\%), {\textcolor{green}{``skateboard''}} (4697, 0.19\%), {\textcolor{blue}{``surfing''}} (4544, 0.18\%), {\textcolor{green}{``water''}} (4513, 0.18\%), {\textcolor{green}{``giraffe''}} (4027, 0.16\%), {\textcolor{green}{``grass''}} (3979, 0.16\%), {\textcolor{green}{``surfboard''}} (3934, 0.16\%), {\textcolor{green}{``wii''}} (3898, 0.16\%), {\textcolor{green}{``kite''}} (3852, 0.16\%), {\textcolor{red}{``10''}} (3756, 0.15\%), {\textcolor{yellow}{``purple''}} (3722, 0.15\%), {\textcolor{green}{``elephant''}} (3646, 0.15\%), {\textcolor{green}{``broccoli''}} (3604, 0.15\%), {\textcolor{green}{``man''}} (3590, 0.14\%), {\textcolor{green}{``winter''}} (3490, 0.14\%), {\textcolor{green}{``stop''}} (3413, 0.14\%), {\textcolor{green}{``train''}} (3226, 0.13\%), {\textcolor{red}{``9''}} (3217, 0.13\%), {\textcolor{green}{``apple''}} (3189, 0.13\%), {\textcolor{green}{``silver''}} (3186, 0.13\%), {\textcolor{green}{``horse''}} (3159, 0.13\%), {\textcolor{green}{``banana''}} (3151, 0.13\%), {\textcolor{green}{``umbrella''}} (3139, 0.13\%), {\textcolor{blue}{``eating''}} (3117, 0.13\%), {\textcolor{green}{``sheep''}} (2927, 0.12\%), {\textcolor{green}{``bear''}} (2803, 0.11\%), {\textcolor{green}{``phone''}} (2772, 0.11\%), {\textcolor{red}{``12''}} (2633, 0.11\%), {\textcolor{green}{``motorcycle''}} (2608, 0.11\%), {\textcolor{green}{``cake''}} (2602, 0.1\%), {\textcolor{green}{``wine''}} (2574, 0.1\%), {\textcolor{green}{``beach''}} (2536, 0.1\%), {\textcolor{green}{``soccer''}} (2504, 0.1\%), {\textcolor{yellow}{``sunny''}} (2475, 0.1\%), {\textcolor{green}{``zebra''}} (2403, 0.1\%), {\textcolor{yellow}{``tan''}} (2402, 0.1\%), {\textcolor{green}{``brick''}} (2395, 0.1\%), {\textcolor{green}{``female''}} (2372, 0.1\%), {\textcolor{green}{``bananas''}} (2350, 0.09\%), {\textcolor{green}{``table''}} (2331, 0.09\%), {\textcolor{green}{``laptop''}} (2316, 0.09\%), {\textcolor{green}{``hat''}} (2277, 0.09\%), {\textcolor{green}{``bench''}} (2259, 0.09\%), {\textcolor{green}{``flowers''}} (2219, 0.09\%), {\textcolor{green}{``woman''}} (2197, 0.09\%), {\textcolor{green}{``male''}} (2170, 0.09\%), {\textcolor{green}{``cow''}} (2084, 0.08\%), {\textcolor{green}{``food''}} (2083, 0.08\%), {\textcolor{green}{``living room''}} (2022, 0.08\%), {\textcolor{green}{``bus''}} (2011, 0.08\%), {\textcolor{green}{``snowboarding''}} (1990, 0.08\%), {\textcolor{green}{``kites''}} (1979, 0.08\%), {\textcolor{green}{``cell phone''}} (1943, 0.08\%), {\textcolor{green}{``helmet''}} (1885, 0.08\%), {\textcolor{cyan}{``maybe''}} (1853, 0.07\%), {\textcolor{cyan}{``outside''}} (1846, 0.07\%), {\textcolor{green}{``hot dog''}} (1809, 0.07\%), {\textcolor{green}{``night''}} (1805, 0.07\%), {\textcolor{green}{``trees''}} (1785, 0.07\%), {\textcolor{red}{``11''}} (1753, 0.07\%), {\textcolor{green}{``bird''}} (1739, 0.07\%), {\textcolor{cyan}{``down''}} (1732, 0.07\%), {\textcolor{green}{``bed''}} (1587, 0.06\%), {\textcolor{green}{``camera''}} (1560, 0.06\%), {\textcolor{green}{``tree''}} (1547, 0.06\%), {\textcolor{green}{``christmas''}} (1544, 0.06\%), {\textcolor{green}{``fence''}} (1543, 0.06\%), {\textcolor{green}{``nothing''}} (1538, 0.06\%), {\textcolor{yellow}{``unknown''}} (1532, 0.06\%), {\textcolor{green}{``tennis racket''}} (1525, 0.06\%), {\textcolor{yellow}{``red and white''}} (1518, 0.06\%), {\textcolor{green}{``bedroom''}} (1500, 0.06\%), {\textcolor{green}{``bat''}} (1494, 0.06\%), {\textcolor{green}{``glasses''}} (1491, 0.06\%), {\textcolor{green}{``tile''}} (1487, 0.06\%), {\textcolor{green}{``metal''}} (1470, 0.06\%), {\textcolor{yellow}{``blue and white''}} (1440, 0.06\%), {\textcolor{green}{``fork''}} (1439, 0.06\%), {\textcolor{green}{``plane''}} (1439, 0.06\%), {\textcolor{green}{``airport''}} (1422, 0.06\%), {\textcolor{yellow}{``cloudy''}} (1413, 0.06\%), {\textcolor{red}{``15''}} (1407, 0.06\%), {\textcolor{cyan}{``up''}} (1399, 0.06\%), {\textcolor{yellow}{``blonde''}} (1398, 0.06\%), {\textcolor{green}{``day''}} (1396, 0.06\%), {\textcolor{green}{``teddy bear''}} (1386, 0.06\%), {\textcolor{green}{``glass''}} (1379, 0.06\%), {\textcolor{red}{``20''}} (1365, 0.05\%), {\textcolor{green}{``beer''}} (1345, 0.05\%), {\textcolor{green}{``car''}} (1331, 0.05\%), {\textcolor{blue}{``sitting''}} (1328, 0.05\%), {\textcolor{green}{``boat''}} (1326, 0.05\%), {\textcolor{blue}{``standing''}} (1326, 0.05\%), {\textcolor{yellow}{``clear''}} (1318, 0.05\%), {\textcolor{red}{``13''}} (1318, 0.05\%), {\textcolor{green}{``nike''}} (1293, 0.05\%), {\textcolor{green}{``sand''}} (1282, 0.05\%), {\textcolor{yellow}{``open''}} (1279, 0.05\%), {\textcolor{green}{``cows''}} (1271, 0.05\%), {\textcolor{green}{``bike''}} (1267, 0.05\%), {\textcolor{green}{``chocolate''}} (1266, 0.05\%), {\textcolor{green}{``donut''}} (1263, 0.05\%), {\textcolor{green}{``airplane''}} (1247, 0.05\%), {\textcolor{green}{``birthday''}} (1241, 0.05\%), {\textcolor{green}{``carrots''}} (1239, 0.05\%), {\textcolor{green}{``skis''}} (1220, 0.05\%), {\textcolor{green}{``girl''}} (1220, 0.05\%), {\textcolor{yellow}{``many''}} (1211, 0.05\%), {\textcolor{green}{``zoo''}} (1204, 0.05\%), {\textcolor{green}{``suitcase''}} (1199, 0.05\%), {\textcolor{yellow}{``old''}} (1180, 0.05\%), {\textcolor{green}{``chair''}} (1174, 0.05\%), {\textcolor{yellow}{``beige''}} (1170, 0.05\%), {\textcolor{green}{``ball''}} (1169, 0.05\%), {\textcolor{green}{``ocean''}} (1168, 0.05\%), {\textcolor{green}{``sandwich''}} (1168, 0.05\%), {\textcolor{green}{``tie''}} (1166, 0.05\%), {\textcolor{green}{``horses''}} (1163, 0.05\%), {\textcolor{green}{``palm''}} (1163, 0.05\%), {\textcolor{green}{``stripes''}} (1155, 0.05\%), {\textcolor{green}{``fall''}} (1146, 0.05\%), {\textcolor{green}{``cheese''}} (1142, 0.05\%), {\textcolor{green}{``scissors''}} (1134, 0.05\%), {\textcolor{green}{``round''}} (1125, 0.05\%), {\textcolor{yellow}{``chinese''}} (1123, 0.05\%), {\textcolor{green}{``knife''}} (1120, 0.05\%), {\textcolor{red}{``14''}} (1110, 0.04\%), {\textcolor{green}{``toilet''}} (1099, 0.04\%), {\textcolor{blue}{``don't know''}} (1085, 0.04\%), {\textcolor{green}{``snowboard''}} (1083, 0.04\%), {\textcolor{green}{``truck''}} (1076, 0.04\%), {\textcolor{green}{``boy''}} (1070, 0.04\%), {\textcolor{green}{``coffee''}} (1070, 0.04\%), {\textcolor{yellow}{``cold''}} (1064, 0.04\%), {\textcolor{green}{``fruit''}} (1064, 0.04\%), {\textcolor{blue}{``walking''}} (1053, 0.04\%), {\textcolor{green}{``wedding''}} (1051, 0.04\%), {\textcolor{green}{``lot''}} (1050, 0.04\%), {\textcolor{green}{``sunglasses''}} (1047, 0.04\%), {\textcolor{green}{``mountains''}} (1030, 0.04\%), {\textcolor{green}{``wall''}} (1009, 0.04\%), {\textcolor{green}{``elephants''}} (1006, 0.04\%), {\textcolor{green}{``wetsuit''}} (998, 0.04\%), {\textcolor{green}{``square''}} (994, 0.04\%), {\textcolor{green}{``toothbrush''}} (989, 0.04\%), {\textcolor{blue}{``sleeping''}} (986, 0.04\%), {\textcolor{green}{``fire hydrant''}} (977, 0.04\%), {\textcolor{green}{``bicycle''}} (973, 0.04\%), {\textcolor{green}{``overcast''}} (968, 0.04\%), {\textcolor{green}{``donuts''}} (961, 0.04\%), {\textcolor{green}{``plastic''}} (961, 0.04\%), {\textcolor{green}{``breakfast''}} (955, 0.04\%), {\textcolor{green}{``tv''}} (953, 0.04\%), {\textcolor{green}{``paper''}} (952, 0.04\%), {\textcolor{green}{``ground''}} (949, 0.04\%), {\textcolor{yellow}{``asian''}} (938, 0.04\%), {\textcolor{green}{``plaid''}} (936, 0.04\%), {\textcolor{green}{``dirt''}} (933, 0.04\%), {\textcolor{green}{``mirror''}} (928, 0.04\%), {\textcolor{green}{``usa''}} (928, 0.04\%), {\textcolor{green}{``chicken''}} (925, 0.04\%), {\textcolor{green}{``plate''}} (920, 0.04\%), {\textcolor{green}{``clock''}} (912, 0.04\%), {\textcolor{green}{``luggage''}} (908, 0.04\%), {\textcolor{green}{``none''}} (908, 0.04\%), {\textcolor{green}{``street''}} (905, 0.04\%), {\textcolor{cyan}{``on table''}} (904, 0.04\%), {\textcolor{green}{``spoon''}} (899, 0.04\%), {\textcolor{blue}{``cooking''}} (898, 0.04\%), {\textcolor{yellow}{``daytime''}} (896, 0.04\%), {\textcolor{red}{``16''}} (893, 0.04\%), {\textcolor{green}{``africa''}} (890, 0.04\%), {\textcolor{green}{``stone''}} (884, 0.04\%), {\textcolor{yellow}{``not sure''}} (873, 0.04\%), {\textcolor{green}{``window''}} (868, 0.03\%), {\textcolor{green}{``sun''}} (865, 0.03\%), {\textcolor{green}{``gold''}} (860, 0.03\%), {\textcolor{green}{``people''}} (856, 0.03\%), {\textcolor{green}{``racket''}} (847, 0.03\%), {\textcolor{green}{``zebras''}} (845, 0.03\%), {\textcolor{green}{``carrot''}} (841, 0.03\%), {\textcolor{green}{``person''}} (835, 0.03\%), {\textcolor{green}{``fish''}} (835, 0.03\%), {\textcolor{yellow}{``happy''}} (824, 0.03\%), {\textcolor{green}{``circle''}} (822, 0.03\%), {\textcolor{green}{``oranges''}} (817, 0.03\%), {\textcolor{green}{``backpack''}} (812, 0.03\%), {\textcolor{red}{``25''}} (810, 0.03\%), {\textcolor{green}{``leaves''}} (809, 0.03\%), {\textcolor{green}{``watch''}} (804, 0.03\%), {\textcolor{green}{``mountain''}} (800, 0.03\%), {\textcolor{green}{``no one''}} (798, 0.03\%), {\textcolor{green}{``ski poles''}} (792, 0.03\%), {\textcolor{green}{``city''}} (791, 0.03\%), {\textcolor{green}{``couch''}} (790, 0.03\%), {\textcolor{green}{``afternoon''}} (782, 0.03\%), {\textcolor{green}{``jeans''}} (781, 0.03\%), {\textcolor{yellow}{``brown and white''}} (779, 0.03\%), {\textcolor{green}{``summer''}} (774, 0.03\%), {\textcolor{green}{``giraffes''}} (772, 0.03\%), {\textcolor{green}{``computer''}} (771, 0.03\%), {\textcolor{green}{``refrigerator''}} (768, 0.03\%), {\textcolor{green}{``birds''}} (762, 0.03\%), {\textcolor{green}{``child''}} (761, 0.03\%), {\textcolor{green}{``park''}} (759, 0.03\%), {\textcolor{blue}{``flying kite''}} (756, 0.03\%), {\textcolor{green}{``restaurant''}} (747, 0.03\%), {\textcolor{green}{``evening''}} (738, 0.03\%), {\textcolor{green}{``graffiti''}} (736, 0.03\%), {\textcolor{red}{``30''}} (730, 0.03\%), {\textcolor{blue}{``grazing''}} (727, 0.03\%), {\textcolor{green}{``flower''}} (723, 0.03\%), {\textcolor{yellow}{``remote''}} (720, 0.03\%), {\textcolor{green}{``hay''}} (719, 0.03\%), {\textcolor{red}{``50''}} (716, 0.03\%).

\section*{Appendix IX: Additional Examples}
\label{sec:dataset_stats}

\begin{figure*}[t]
\centering
\includegraphics[width=1\linewidth]{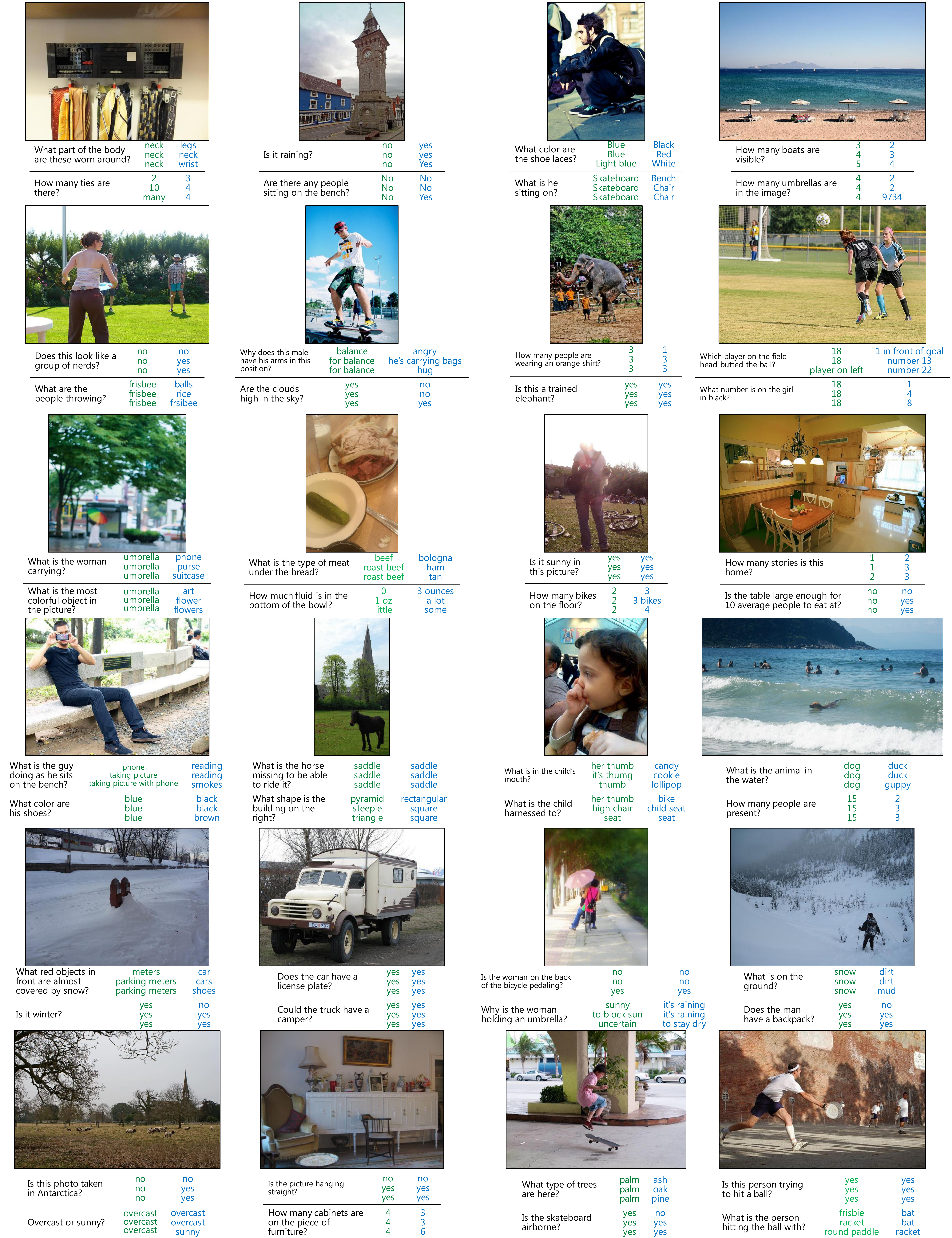}
\caption{Random examples of questions (black), \change{(a subset of the)} answers given when looking at the image (green), and answers given when not looking at the image (blue) for numerous representative examples of the real image dataset.}
\label{fig:coco_more_examples}
\end{figure*}

\begin{figure*}[t]
\centering
\includegraphics[width=1\linewidth]{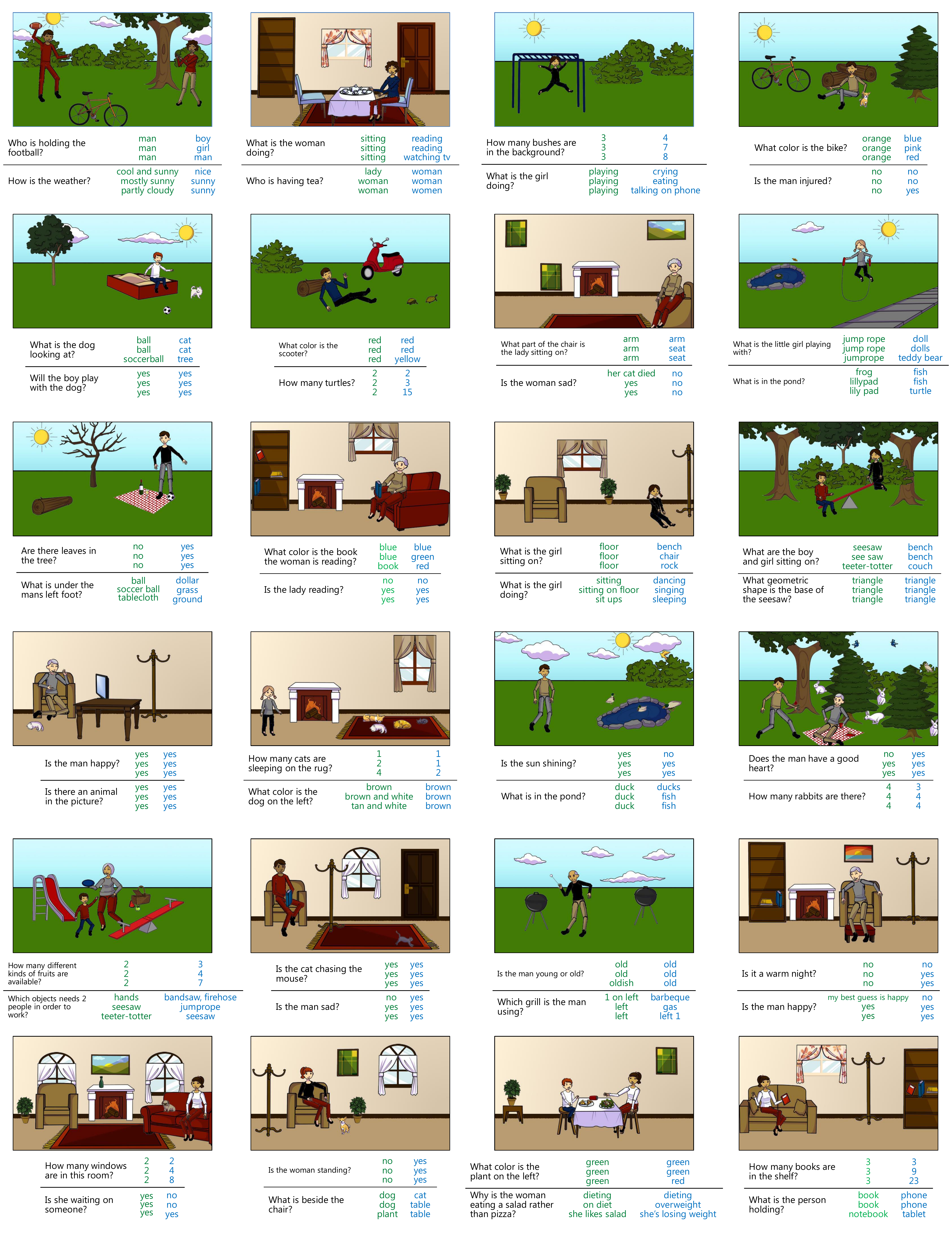}
\caption{Random examples of questions (black), \change{(a subset of the)} answers given when looking at the image (green), and answers given when not looking at the image (blue) for numerous representative examples of the abstract scene dataset.}
\label{fig:abstract_more_examples}
\end{figure*}

\begin{figure*}[t]
\centering
\includegraphics[width=1\linewidth]{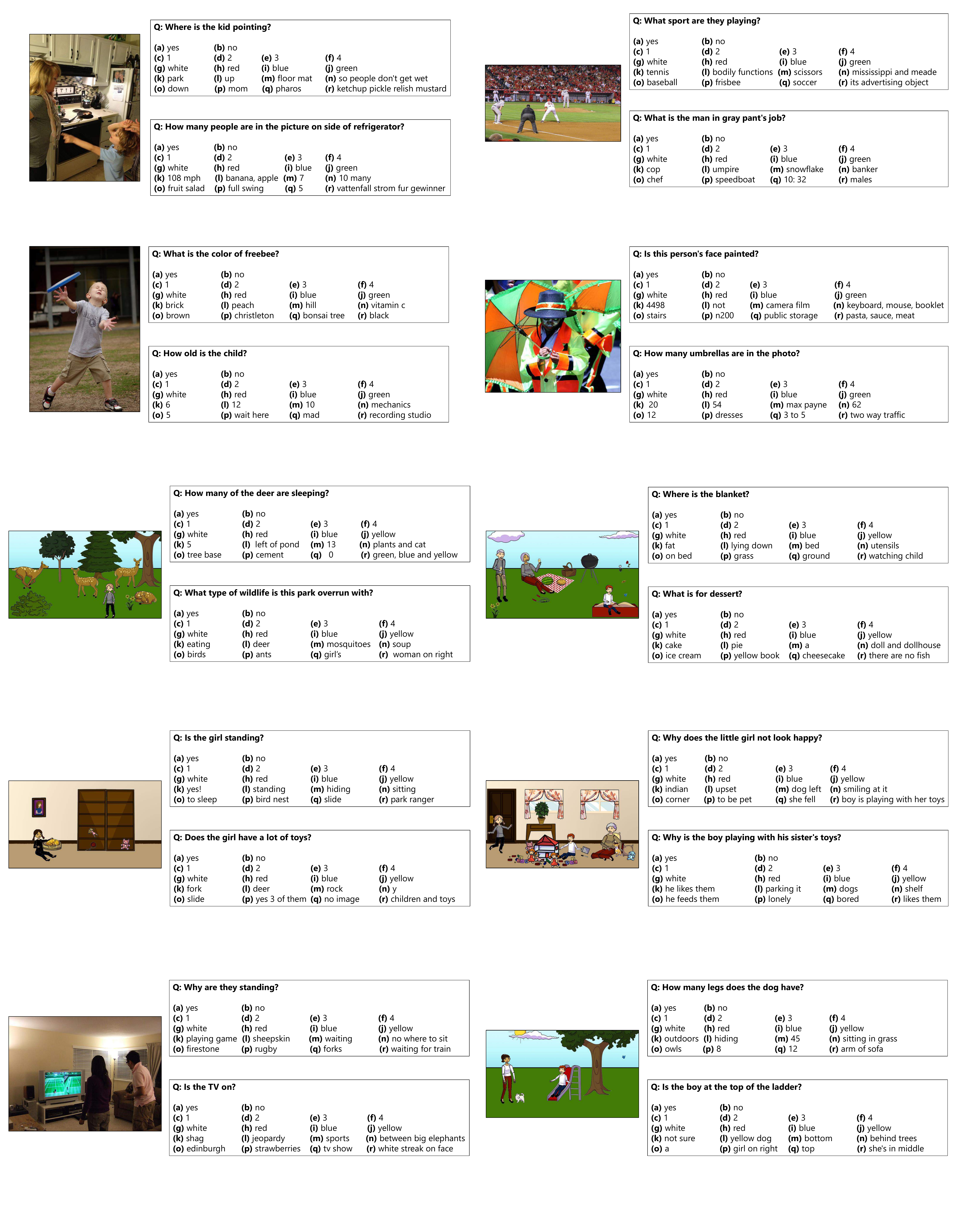}
\caption{Random examples of multiple-choice questions for numerous representative examples of the real and abstract scene dataset.}
\label{fig:mc_examples}
\end{figure*}

To provide insight into the dataset, we provide additional examples. In \figref{fig:coco_more_examples}, \figref{fig:abstract_more_examples}, and \figref{fig:mc_examples}, we show a random selection of the VQA dataset for the MS COCO~\cite{coco} images, abstract scenes, and multiple-choice questions, respectively.

%% file: vqa_main.bbl
\begin{thebibliography}{10}\itemsep=-1pt

\bibitem{cloudcv}
H.~Agrawal, C.~S. Mathialagan, Y.~Goyal, N.~Chavali, P.~Banik, A.~Mohapatra,
  A.~Osman, and D.~Batra.
\newblock Cloudcv: Large-scale distributed computer vision as a cloud service.
\newblock In {\em Mobile Cloud Visual Media Computing}, pages 265--290.
  Springer International Publishing, 2015.

\bibitem{Antol2014}
S.~Antol, C.~L. Zitnick, and D.~Parikh.
\newblock {Zero-Shot Learning via Visual Abstraction}.
\newblock In {\em ECCV}, 2014.

\bibitem{vizwiz}
J.~P. Bigham, C.~Jayant, H.~Ji, G.~Little, A.~Miller, R.~C. Miller, R.~Miller,
  A.~Tatarowicz, B.~White, S.~White, and T.~Yeh.
\newblock {VizWiz: Nearly Real-time Answers to Visual Questions}.
\newblock In {\em {User Interface Software and Technology}}, 2010.

\bibitem{Freebase}
K.~Bollacker, C.~Evans, P.~Paritosh, T.~Sturge, and J.~Taylor.
\newblock {Freebase: A Collaboratively Created Graph Database for Structuring
  Human Knowledge}.
\newblock In {\em International Conference on Management of Data}, 2008.

\bibitem{NEEL}
A.~Carlson, J.~Betteridge, B.~Kisiel, B.~Settles, E.~R.~H. Jr., and T.~M.
  Mitchell.
\newblock {Toward an Architecture for Never-Ending Language Learning}.
\newblock In {\em AAAI}, 2010.

\bibitem{DBLP:journals/corr/ChenFLVGDZ15}
X.~Chen, H.~Fang, T.~Lin, R.~Vedantam, S.~Gupta, P.~Doll{\'{a}}r, and C.~L.
  Zitnick.
\newblock Microsoft {COCO} captions: Data collection and evaluation server.
\newblock {\em CoRR}, abs/1504.00325, 2015.

\bibitem{capeval2015}
X.~Chen, H.~Fang, T.-Y. Lin, R.~Vedantam, S.~Gupta, P.~Doll{\'{a}}r, and C.~L.
  Zitnick.
\newblock {Microsoft COCO Captions: Data Collection and Evaluation Server}.
\newblock {\em arXiv preprint arXiv:1504.00325}, 2015.

\bibitem{NEIL}
X.~Chen, A.~Shrivastava, and A.~Gupta.
\newblock {NEIL: Extracting Visual Knowledge from Web Data}.
\newblock In {\em ICCV}, 2013.

\bibitem{captioning_xinlei}
X.~Chen and C.~L. Zitnick.
\newblock {Mind's Eye: A Recurrent Visual Representation for Image Caption
  Generation}.
\newblock In {\em CVPR}, 2015.

\bibitem{CoppersmithKelly14}
G.~Coppersmith and E.~Kelly.
\newblock Dynamic wordclouds and vennclouds for exploratory data analysis.
\newblock In {\em ACL Workshop on Interactive Language Learning and
  Visualization}, 2014.

\bibitem{deng}
J.~Deng, A.~C. Berg, and L.~Fei-Fei.
\newblock {Hierarchical Semantic Indexing for Large Scale Image Retrieval}.
\newblock In {\em CVPR}, 2011.

\bibitem{captioning_berkeley}
J.~Donahue, L.~A. Hendricks, S.~Guadarrama, M.~Rohrbach, S.~Venugopalan,
  K.~Saenko, and T.~Darrell.
\newblock {Long-term Recurrent Convolutional Networks for Visual Recognition
  and Description}.
\newblock In {\em CVPR}, 2015.

\bibitem{elliott2014comparing}
D.~Elliott and F.~Keller.
\newblock {Comparing Automatic Evaluation Measures for Image Description}.
\newblock In {\em ACL}, 2014.

\bibitem{zettleymoyer_acl13}
A.~Fader, L.~Zettlemoyer, and O.~Etzioni.
\newblock {Paraphrase-Driven Learning for Open Question Answering}.
\newblock In {\em ACL}, 2013.

\bibitem{zettlemoyer_kdd14}
A.~Fader, L.~Zettlemoyer, and O.~Etzioni.
\newblock {Open Question Answering over Curated and Extracted Knowledge Bases}.
\newblock In {\em International Conference on Knowledge Discovery and Data
  Mining}, 2014.

\bibitem{captioning_msr}
H.~Fang, S.~Gupta, F.~N. Iandola, R.~Srivastava, L.~Deng, P.~Doll{\'{a}}r,
  J.~Gao, X.~He, M.~Mitchell, J.~C. Platt, C.~L. Zitnick, and G.~Zweig.
\newblock {From Captions to Visual Concepts and Back}.
\newblock In {\em CVPR}, 2015.

\bibitem{FarhadiSentencesECCV2010}
A.~Farhadi, M.~Hejrati, A.~Sadeghi, P.~Young, C.~Rashtchian, J.~Hockenmaier,
  and D.~Forsyth.
\newblock {Every Picture Tells a Story: Generating Sentences for Images}.
\newblock In {\em ECCV}, 2010.

\bibitem{baiduVQA}
H.~Gao, J.~Mao, J.~Zhou, Z.~Huang, and A.~Yuille.
\newblock Are you talking to a machine? dataset and methods for multilingual
  image question answering.
\newblock In {\em NIPS}, 2015.

\bibitem{geman}
D.~Geman, S.~Geman, N.~Hallonquist, and L.~Younes.
\newblock {A Visual Turing Test for Computer Vision Systems}.
\newblock In {\em PNAS}, 2014.

\bibitem{GordonVanDurme13}
J.~Gordon and B.~V. Durme.
\newblock Reporting bias and knowledge extraction.
\newblock In {\em Proceedings of the 3rd Workshop on Knowledge Extraction, at
  CIKM 2013}, 2013.

\bibitem{youtube2text}
S.~Guadarrama, N.~Krishnamoorthy, G.~Malkarnenkar, S.~Venugopalan, R.~Mooney,
  T.~Darrell, and K.~Saenko.
\newblock {YouTube2Text: Recognizing and Describing Arbitrary Activities Using
  Semantic Hierarchies and Zero-Shot Recognition}.
\newblock In {\em ICCV}, December 2013.

\bibitem{hodosh2013framing}
M.~Hodosh, P.~Young, and J.~Hockenmaier.
\newblock {Framing Image Description as a Ranking Task: Data, Models and
  Evaluation Metrics}.
\newblock {\em JAIR}, 2013.

\bibitem{jia2014caffe}
Y.~Jia, E.~Shelhamer, J.~Donahue, S.~Karayev, J.~Long, R.~Girshick,
  S.~Guadarrama, and T.~Darrell.
\newblock Caffe: Convolutional architecture for fast feature embedding.
\newblock {\em arXiv preprint arXiv:1408.5093}, 2014.

\bibitem{captioning_stanford}
A.~Karpathy and L.~Fei{-}Fei.
\newblock {Deep Visual-Semantic Alignments for Generating Image Descriptions}.
\newblock In {\em CVPR}, 2015.

\bibitem{referit}
S.~Kazemzadeh, V.~Ordonez, M.~Matten, and T.~L. Berg.
\newblock {ReferItGame: Referring to Objects in Photographs of Natural Scenes}.
\newblock In {\em EMNLP}, 2014.

\bibitem{captioning_toronto}
R.~Kiros, R.~Salakhutdinov, and R.~S. Zemel.
\newblock {Unifying Visual-Semantic Embeddings with Multimodal Neural Language
  Models}.
\newblock {\em TACL}, 2015.

\bibitem{kiros2015skip}
R.~Kiros, Y.~Zhu, R.~Salakhutdinov, R.~S. Zemel, A.~Torralba, R.~Urtasun, and
  S.~Fidler.
\newblock Skip-thought vectors.
\newblock {\em arXiv preprint arXiv:1506.06726}, 2015.

\bibitem{nounCoref2014}
C.~Kong, D.~Lin, M.~Bansal, R.~Urtasun, and S.~Fidler.
\newblock {What Are You Talking About? Text-to-Image Coreference}.
\newblock In {\em CVPR}, 2014.

\bibitem{AlexNet}
A.~Krizhevsky, I.~Sutskever, and G.~E. Hinton.
\newblock {ImageNet Classification with Deep Convolutional Neural Networks}.
\newblock In {\em NIPS}, 2012.

\bibitem{babytalk}
G.~Kulkarni, V.~Premraj, S.~L. Sagnik Dhar~and, Y.~Choi, A.~C. Berg, and T.~L.
  Berg.
\newblock {Baby {T}alk: Understanding and Generating Simple Image
  Descriptions}.
\newblock In {\em CVPR}, 2011.

\bibitem{Cyc}
D.~B. Lenat and R.~V. Guha.
\newblock {\em Building Large Knowledge-Based Systems; Representation and
  Inference in the Cyc Project}.
\newblock Addison-Wesley Longman Publishing Co., Inc., 1989.

\bibitem{coco}
T.-Y. Lin, M.~Maire, S.~Belongie, J.~Hays, P.~Perona, D.~Ramanan,
  P.~Doll{\'{a}}r, and C.~L. Zitnick.
\newblock {Microsoft COCO: Common Objects in Context}.
\newblock In {\em ECCV}, 2014.

\bibitem{FITB_VP}
X.~Lin and D.~Parikh.
\newblock {Don't Just Listen, Use Your Imagination: Leveraging Visual Common
  Sense for Non-Visual Tasks}.
\newblock In {\em CVPR}, 2015.

\bibitem{Lin_2015_CVPR}
X.~Lin and D.~Parikh.
\newblock Don't just listen, use your imagination: Leveraging visual common
  sense for non-visual tasks.
\newblock In {\em CVPR}, 2015.

\bibitem{ConceptNet}
H.~Liu and P.~Singh.
\newblock {ConceptNet --- A Practical Commonsense Reasoning Tool-Kit}.
\newblock {\em BT Technology Journal}, 2004.

\bibitem{fritz}
M.~Malinowski and M.~Fritz.
\newblock {A Multi-World Approach to Question Answering about Real-World Scenes
  based on Uncertain Input}.
\newblock In {\em NIPS}, 2014.

\bibitem{Malinowski_2015_ICCV}
M.~Malinowski, M.~Rohrbach, and M.~Fritz.
\newblock Ask your neurons: A neural-based approach to answering questions
  about images.
\newblock In {\em ICCV}, 2015.

\bibitem{captioning_baidu_ucla}
J.~Mao, W.~Xu, Y.~Yang, J.~Wang, and A.~L. Yuille.
\newblock {Explain Images with Multimodal Recurrent Neural Networks}.
\newblock {\em CoRR}, abs/1410.1090, 2014.

\bibitem{word2vec}
T.~Mikolov, I.~Sutskever, K.~Chen, G.~S. Corrado, and J.~Dean.
\newblock {Distributed Representations of Words and Phrases and their
  Compositionality}.
\newblock In {\em NIPS}, 2013.

\bibitem{MitchellEtAl12}
M.~Mitchell, J.~Dodge, A.~Goyal, K.~Yamaguchi, K.~Stratos, X.~Han, A.~Mensch,
  A.~Berg, T.~L. Berg, and H.~Daume~III.
\newblock {Midge: Generating Image Descriptions From Computer Vision
  Detections}.
\newblock In {\em ACL}, 2012.

\bibitem{MitchellEtAl13}
M.~Mitchell, K.~van Deemter, and E.~Reiter.
\newblock Attributes in visual reference.
\newblock In {\em PRE-CogSci}, 2013.

\bibitem{mitchell2013generating}
M.~Mitchell, K.~Van~Deemter, and E.~Reiter.
\newblock {Generating Expressions that Refer to Visible Objects}.
\newblock In {\em HLT-NAACL}, 2013.

\bibitem{peopleCoref2014}
V.~Ramanathan, A.~Joulin, P.~Liang, and L.~Fei-Fei.
\newblock {Linking People with ``Their" Names using Coreference Resolution}.
\newblock In {\em ECCV}, 2014.

\bibitem{Ren_2015_NIPS}
M.~Ren, R.~Kiros, and R.~Zemel.
\newblock Exploring models and data for image question answering.
\newblock In {\em NIPS}, 2015.

\bibitem{richardson2013mctest}
M.~Richardson, C.~J. Burges, and E.~Renshaw.
\newblock {MCTest: A Challenge Dataset for the Open-Domain Machine
  Comprehension of Text}.
\newblock In {\em EMNLP}, 2013.

\bibitem{video}
M.~Rohrbach, W.~Qiu, I.~Titov, S.~Thater, M.~Pinkal, and B.~Schiele.
\newblock {Translating Video Content to Natural Language Descriptions}.
\newblock In {\em ICCV}, 2013.

\bibitem{Sadeghi_2015_CVPR}
F.~Sadeghi, S.~K. Kumar~Divvala, and A.~Farhadi.
\newblock Viske: Visual knowledge extraction and question answering by visual
  verification of relation phrases.
\newblock In {\em CVPR}, 2015.

\bibitem{Simonyan14c}
K.~Simonyan and A.~Zisserman.
\newblock Very deep convolutional networks for large-scale image recognition.
\newblock {\em CoRR}, abs/1409.1556, 2014.

\bibitem{StanfordPOS}
K.~Toutanova, D.~Klein, C.~D. Manning, and Y.~Singer.
\newblock Feature-rich part-of-speech tagging with a cyclic dependency network.
\newblock In {\em ACL}, 2003.

\bibitem{SongChun_video_queries}
K.~Tu, M.~Meng, M.~W. Lee, T.~E. Choe, and S.~C. Zhu.
\newblock {Joint Video and Text Parsing for Understanding Events and Answering
  Queries}.
\newblock {\em {IEEE} MultiMedia}, 2014.

\bibitem{cider}
R.~Vedantam, C.~L. Zitnick, and D.~Parikh.
\newblock {CIDEr}: {C}onsensus-based {I}mage {D}escription {E}valuation.
\newblock In {\em CVPR}, 2015.

\bibitem{Vendantam_2015_ICCV}
R.~Vendantam, X.~Lin, T.~Batra, C.~L. Zitnick, and D.~Parikh.
\newblock Learning common sense through visual abstraction.
\newblock In {\em ICCV}, 2015.

\bibitem{captioning_google}
O.~Vinyals, A.~Toshev, S.~Bengio, and D.~Erhan.
\newblock {Show and Tell: {A} Neural Image Caption Generator}.
\newblock In {\em CVPR}, 2015.

\bibitem{weston_qa}
J.~Weston, A.~Bordes, S.~Chopra, and T.~Mikolov.
\newblock {Towards AI-Complete Question Answering: A Set of Prerequisite Toy
  Tasks}.
\newblock {\em CoRR}, abs/1502.05698, 2015.

\bibitem{Yu_2015_ICCV}
L.~Yu, E.~Park, A.~C. Berg, and T.~L. Berg.
\newblock Visual madlibs: Fill-in-the-blank description generation and question
  answering.
\newblock In {\em ICCV}, 2015.

\bibitem{yinyang}
P.~Zhang, Y.~Goyal, D.~Summers{-}Stay, D.~Batra, and D.~Parikh.
\newblock Yin and yang: Balancing and answering binary visual questions.
\newblock {\em CoRR}, abs/1511.05099, 2015.

\bibitem{ZitnickCVPR2013}
C.~L. Zitnick and D.~Parikh.
\newblock {Bringing Semantics Into Focus Using Visual Abstraction}.
\newblock In {\em CVPR}, 2013.

\bibitem{ZitnickICCV2013}
C.~L. Zitnick, D.~Parikh, and L.~Vanderwende.
\newblock {Learning the Visual Interpretation of Sentences}.
\newblock In {\em ICCV}, 2013.

\bibitem{VedantamPAMI2015}
C.~L. Zitnick, R.~Vedantam, and D.~Parikh.
\newblock {Adopting Abstract Images for Semantic Scene Understanding}.
\newblock {\em PAMI}, 2015.

\end{thebibliography}
